%% file: main.tex
\documentclass[10pt,twocolumn,letterpaper]{article}

\usepackage{wacv}
\usepackage{times}
\usepackage{epsfig}
\usepackage{graphicx}
\usepackage{amsmath}
\usepackage{amssymb}

\usepackage{subcaption}
\usepackage{lipsum}
\usepackage{color,soul}
\usepackage{booktabs}
\usepackage{mwe}
\usepackage{multicol,multirow} 
\usepackage{wrapfig}
\usepackage{cite}

\def\wacvPaperID{1241} % Enter the WACV Paper ID here

\wacvfinalcopy % *** Uncomment this line for the final submission

\ifwacvfinal
\def\assignedStartPage{9876} % *** Enter the assigned starting page number (instead of 9876)
\fi
\ifwacvfinal
\usepackage[breaklinks=true,bookmarks=false]{hyperref}
\else
\usepackage[pagebackref=true,breaklinks=true,colorlinks,bookmarks=false]{hyperref}
\fi
\usepackage[export]{adjustbox}

\usepackage{cleveref}

\ifwacvfinal
\else
\fi

\begin{document}

%%%%%%%%% TITLE
\title{Class-agnostic Object Detection}

\author{Ayush Jaiswal, Yue Wu, Pradeep Natarajan, Premkumar Natarajan\\
Amazon Alexa Natural Understanding\\
Manhattan Beach, CA, USA\\
{\tt\small \{ayujaisw, wuayue, natarap, premknat\}@amazon.com}
% For a paper whose authors are all at the same institution,
% omit the following lines up until the closing ``}''.
% Additional authors and addresses can be added with ``\and'',
% just like the second author.
% To save space, use either the email address or home page, not both
}

\twocolumn[{%
\renewcommand\twocolumn[1][]{#1}%
\maketitle
\begin{center}
    \centering
    \vspace{-15pt}
    \includegraphics[trim={0 0 0 0.06cm},clip,width=\textwidth]{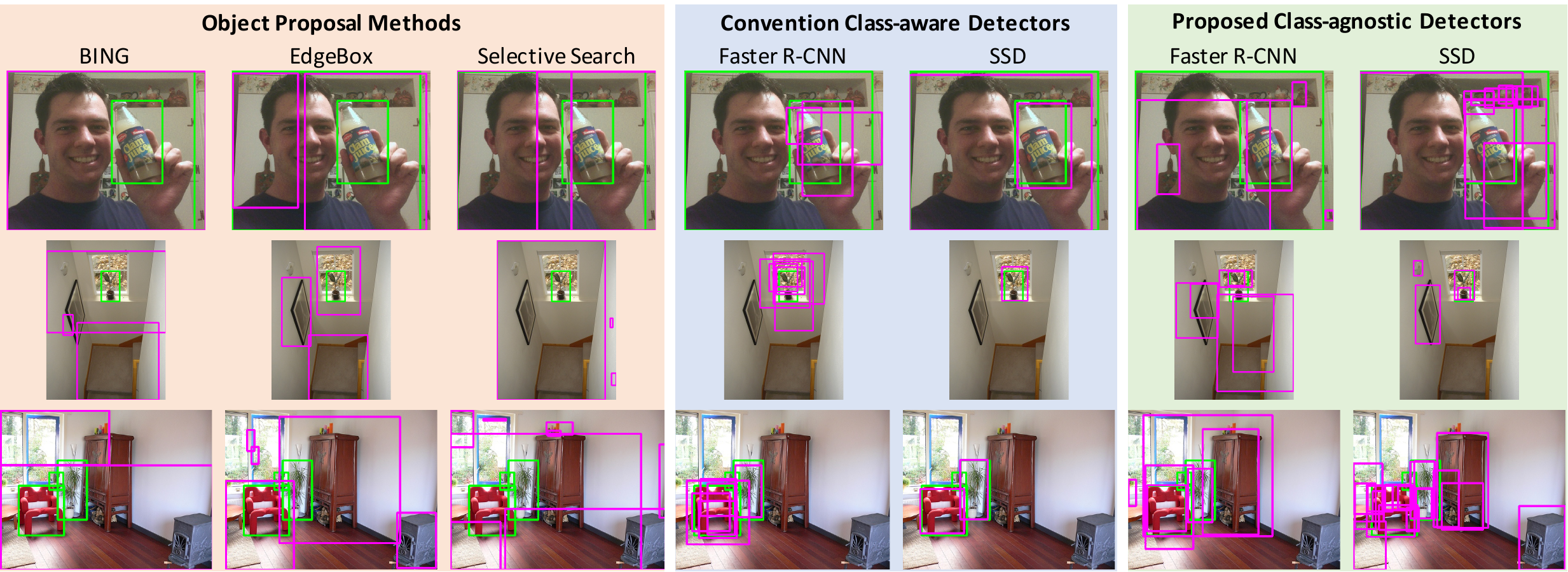}
    \captionof{figure}{\label{fig:teaser}Class-agnostic object detection using object proposal methods (OPMs), conventional class-aware detectors, and the proposed adversarially trained class-agnostic detectors. \textcolor{green}{Green} and \textcolor{magenta}{magenta} boxes indicate \textcolor{green}{ground-truth} and \textcolor{magenta}{detection outputs}, respectively. OPMs generate regions of interest that could contain objects without precisely locating any object. Class-aware detectors detect objects from a set of \emph{known} object-types. The proposed class-agnostic detection aims to localize all objects irrespective of their types including those of unknown classes. Results show that the proposed adversarially trained class-agnostic models detect novel objects for which no annotations are available. Best viewed digitally and zoomed in.}%
\end{center}%
}]

% \maketitle
%\thispagestyle{empty}

%%%%%%%%% ABSTRACT
\begin{abstract}
    \vspace{-5pt}
   Object detection models perform well at localizing and classifying objects that they are shown during training. However, due to the difficulty and cost associated with creating and annotating detection datasets, trained models detect a limited number of object types with unknown objects treated as background content. This hinders the adoption of conventional detectors in real-world applications like large-scale object matching, visual grounding, visual relation prediction, obstacle detection (where it is more important to determine the presence and location of objects than to find specific types), etc. We propose class-agnostic object detection as a new problem that focuses on detecting objects irrespective of their object-classes. Specifically, the goal is to predict bounding boxes for all objects in an image but not their object-classes. The predicted boxes can then be consumed by another system to perform application-specific classification, retrieval, etc. We propose training and evaluation protocols for benchmarking class-agnostic detectors to advance future research in this domain. Finally, we propose (1) baseline methods and (2) a new adversarial learning framework for class-agnostic detection that forces the model to exclude class-specific information from features used for predictions. Experimental results show that adversarial learning improves class-agnostic detection efficacy.
\end{abstract}

%%%%%%%%% SECTIONS

\input{sections/01_introduction}
\input{sections/02_related_work}
\input{sections/03_caod}
\input{sections/04_method}
\input{sections/05_evaluation}
\input{sections/06_conclusion}

{\small
\bibliographystyle{ieee_fullname}
\bibliography{main}
}

\end{document}

%% file: sections/01_introduction.tex
\vspace{-20pt}
\section{Introduction}
\label{sec:introduction}

Human visual scene understanding relies on the ability to detect and recognize objects in one's field of view. This naturally carries over to machine scene understanding through computer vision techniques. Hence, the field of object detection has garnered tremendous research interest~\cite{bib:obj_survey} and large advances have been made in improving detection systems, especially since the adoption of deep learning.

The object detection task is formulated as the prediction of bounding boxes and classes for objects in a given image. This requires densely labeled data that contains annotations for all objects in training images. However, creating such datasets is extremely challenging and expensive. Therefore, conventional object detection focuses instead on the reduced task of locating and recognizing ``objects of interest'' corresponding to limited types of objects that are labeled in the training data, with objects of unknown types treated as background content. Notably, detection of unknown object-types is explicitly penalized in the widely adopted mean average precision metric used for benchmarking. This hinders the adoption of trained detectors in real-world applications due to the added cost of retraining them for application-specific object types. Furthermore, such detectors cannot be used in applications like obstacle detection, where it is more important to determine the location of all objects present in the scene than to find specific kinds of objects.

In order to address the aforementioned limitations of conventional class-aware detection, we propose class-agnostic object detection as a new research task with the goal of predicting bounding boxes for all objects present in an image irrespective of their object-types. Intuitively, this task additionally seeks to detect objects of unknown types, which are not present or annotated in the training data. This challenging yet high-reward goal of generalization to unseen object types would benefit downstream applications (e.g., application-specific object classification, object retrieval from large databases, etc.) that can consume such class-agnostic detections.

Besides the problem formulation, we propose training and evaluation protocols for class-agnostic detection with generalization and downstream utility as primary goals. Generalization is evaluated on PASCAL VOC~\cite{bib:voc}, MS COCO~\cite{bib:coco}, and Open Images~\cite{bib:oi} in the form of recall of unseen classes. Specifically, models trained on VOC are evaluated on unseen classes of COCO and those trained on COCO are tested on non-overlapping classes of Open Images. Furthermore, the VOC dataset is split into seen and unseen classes in order to measure recall of seen and unseen classes within the same dataset. Utility of trained detectors is evaluated as the accuracy of pretrained ImageNet~\cite{bib:imagenet} classifiers on ObjectNet~\cite{bib:objectnet} images cropped using the bounding boxes predicted by the detectors.

We present a few baseline methods for class-agnostic object detection -- region proposals of two-stage detectors, pretrained class-aware models used \textit{as-is} or finetuned for binary object versus not classification, and detection models trained from scratch for the said binary classification. Finally, we propose a new adversarial training framework that forces the detection model to exclude class-specific information from the features used for making predictions. Experimental results show that our adversarial framework improves class-agnostic detection efficacy for both Faster R-CNN~\cite{bib:frcnn} (two-stage) and SSD~\cite{bib:ssd} (one-stage).

The major contributions of this paper are:
\vspace{-5pt}
\begin{itemize}
    \setlength\itemsep{-0.5em}
    \item a novel class-agnostic object detection problem formulation as a new research direction,
    \item training and evaluation protocols for benchmarking and advancing research,
    \item a new adversarial learning framework for class-agnostic detection that penalizes the model if object-type is encoded in embeddings used for predictions.
\end{itemize}

%% file: sections/02_related_work.tex
\section{Related Work}
\label{sec:related_work}

Zou \textit{et al.}~\cite{bib:obj_survey} provide a comprehensive survey of advances in object detection in the past 20 years. Some conventional detection works~\cite{bib:gen_obj_1,bib:gen_obj_2,bib:gen_obj_3,bib:gen_obj_4,bib:gen_obj_5} describe the task as ``generic'' object detection but they use it to signify their focus on common objects of \emph{known} types, which is distinct from our work on detecting objects of \emph{all} types.

Two-stage object detectors like Faster R-CNN~\cite{bib:frcnn} employ a region proposal module that identifies a set of regions in a given image that could contain objects, which are then used by a downstream detector module to localize and identify objects. A number of works~\cite{bib:obj_proposal_1,bib:obj_proposal_2,bib:obj_proposal_3,bib:objectness_2,bib:selective_search,bib:obj_prop_pont,bib:obj_prop_endres,bib:obj_prop_zhang} have been proposed recently that aim to improve the quality of such \textit{object proposals} and reduce their number in order to speed up the final detection task. A majority of proposal methods are trained end-to-end with the detector module, which biases them to objects of known types. Methods also exist for generating~\cite{bib:edge_proposal_1,bib:edge_proposal_2_bing,bib:edge_proposal_3_edgeboxes,bib:edge_proposal_4} or filtering~\cite{bib:filter_proposal_1,bib:filter_proposal_2,bib:filter_proposal_3} object proposals based on edge-related features and objectness metrics. However, as defined, these proposals do not correspond directly to final detections and must be fed to a detector module to infer final detections. In contrast, the proposed task of class-agnostic detection is aimed at predicting final bounding boxes for all objects in an image. Two-stage methods for class-agnostic detection could employ these works to generate intermediate object proposals.

Some works~\cite{bib:objectness_1,bib:pixel_objectness} focus on generating pixel-level objectness scores in order to segment object regions out of background content. These works produce objectness heatmaps which could sometimes be used to detect individual objects if they do not overlap and are strikingly distinct.

A few works~\cite{bib:class_generic,bib:scalable} have shown the efficacy of convolutional neural networks like AlexNet~\cite{bib:alexnet} for localizing objects irrespective of their classes. In this work, we take this idea further and formally define the task of class-agnostic object detection along with training and evaluation protocols for benchmarking current and future research.

Kuen \textit{et al.}~\cite{bib:scaling_obj_det} show how weights from object recognition networks can be transferred to detection models to scale-up the number of object-types for detection. This line of research is also related to few-shot~\cite{bib:incremental_obj_det,bib:fsd_1,bib:fsd_2,bib:fsd_3,bib:fsd_4,bib:fsd_5,bib:fsd_6} and zero-shot~\cite{bib:zsd_1,bib:zsd_2,bib:zsd_3} object detection, which are targeted towards detecting objects of novel types given a few reference samples or descriptions, respectively.

%% file: sections/03_caod.tex
\begin{figure*}
    \centering
    \includegraphics[trim={0 0.1cm 0 0},clip,width=0.75\textwidth]{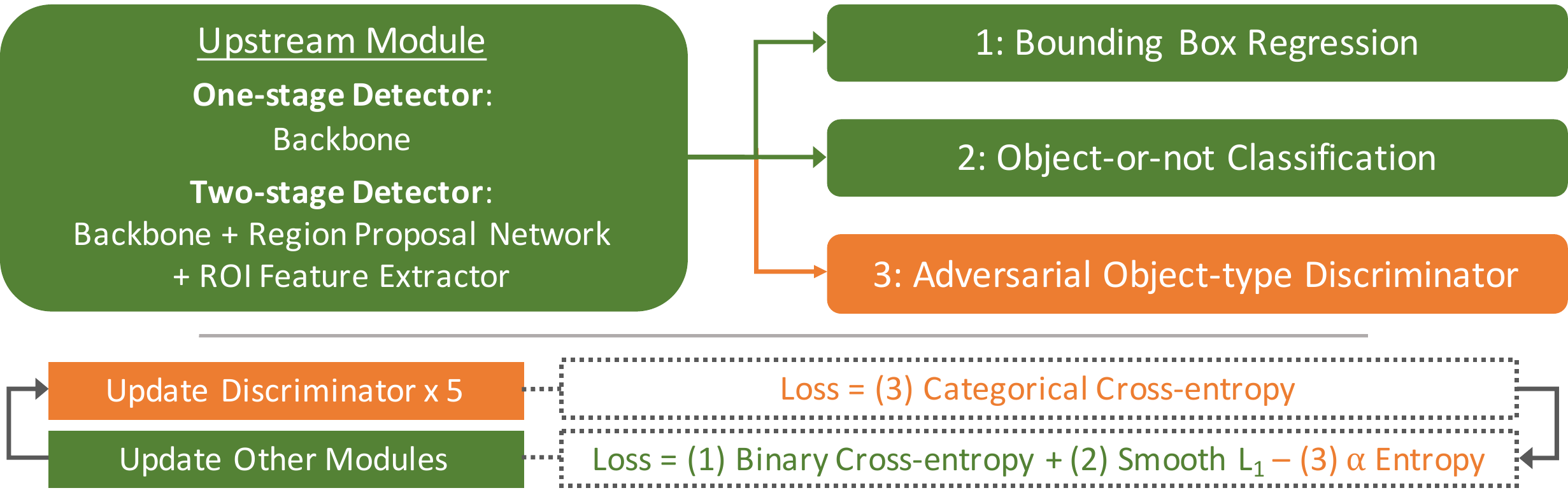}
    \caption{Adversarial framework for class-agnostic object detection for both one- and two-stage models. The upstream module with the backbone (e.g., VGG16, ResNet-50, etc.), Region Proposal Network (two-stage) and ROI feature extractor (two-stage) remain unchanged, along with the box regressor. The conventional object-type classification is replaced with a binary object-or-not classifier, and a new adversarial object-type discriminator is attached during training. The bottom part of the figure shows the training procedure -- iterating between five discriminator updates for each update to the other modules.}
    \label{fig:model}
    \vspace{-5pt}
\end{figure*}

\section{Class-agnostic Object Detection}
\label{sec:caod}

In this section, we describe the proposed task of class-agnostic object detection and contrast it with conventional class-aware object detection. We then discuss a few intuitive baseline models for the proposed task.

\subsection{From Class-aware to Class-agnostic Detection}

The conventional formulation of object detection can be described as the task of locating and classifying objects in a given image. The goal of models developed to solve this task is to predict bounding box coordinates and class labels for objects present in the image. Given this formulation, a plethora of models have been proposed that train on images with object annotations. However, due to the difficulty and cost associated with collecting and labeling large datasets, annotations are typically collected for a limited number of object categories. This has inadvertently reduced the original formulation to the task of detecting ``objects of interest'', corresponding to object types that are annotated in training datasets. Although reduced in form, the conventional formulation remains a challenging problem, with direct application in use-cases with a fixed set of known object types.

In this work, we propose a sibling task of class-agnostic object detection that aims to detect all objects in a given image irrespective of their object-types. More specifically, the goal of this task is to predict bounding boxes for all objects present in an image but not their category. Furthermore, given that most available training datasets do not contain dense annotations for all kinds of objects, an additional implicit goal for models developed for this task is to generalize to objects of unknown types, i.e., those for which annotations are not available in the training dataset. This is in direct contrast to conventional class-aware object detection, which treats unknown objects as background content. As compared to conventional detectors, class-agnostic models can be more easily adopted in complex real-life applications like object retrieval from a large and diverse database, recognition of application-specific object-types through training a downstream object recognition model (instead of retraining the full detection network), etc.

\subsection{Baseline Models}

We identify three intuitive and straightforward baseline models for solving the class-agnostic detection task, selected due to their ease of implementation and natural curiosity about their performance. The baseline models are:
\begin{itemize}
    \setlength\itemsep{-0.4em}
    \item region proposal network of a two-stage detector,
    \item class-aware detector trained for object-type classification and bounding box regression,
    \item pretrained class-aware model finetuned end-to-end for object-or-not binary classification instead of object-type, along with bounding box regression,
    \item detection model trained from scratch for object-or-not binary classification and bounding box regression.
\end{itemize}

%% file: sections/04_method.tex
\section{Adversarial Learning Framework for \\Class-agnostic Object Detection}
\label{sec:method}

\subsection{General Framework}

Conventional class-aware detection focuses on detecting ``objects of interest'', which inherently requires models to be able to distinguish between types of objects in a closed known set. Intuitively, models achieve this by encoding features that are discriminative of object-types. However, for class-agnostic detection and for models to be able to detect previously unseen types of objects, detectors should encode features that more effectively distinguish objects from background content and individual objects from other objects in the image, without discriminating between object-types.

Na\"ively training conventional object detectors for the binary classification task of object-or-not along with bounding box regression is not sufficient to ensure that models focus on class-agnostic features and more importantly, ignore type-distinguishing features so that they can better generalize to unseen object-types. In order to overcome this problem, we propose to train class-agnostic object detectors in an adversarial fashion such that models are penalized for encoding features that contain object-type information.

We begin with observing a popular two-part pattern in the model design of both one-stage and two-stage conventional object detectors. The first \textit{upstream} part of a detection model learns a set of convolutional features from entire images (one-stage) or regions of interest (two-stage). The second \textit{downstream} part consumes these features and passes them through classification and regression branches for object-type and bounding box prediction, respectively. This two-part setup allows for external control on the information output by the first part and consumed by the second.

We propose to augment class-agnostic detectors with adversarial discriminator branches that attempt to classify object-types (annotated in the training data) from the features output by the upstream part of detection networks, and penalize the model training if they are successful~\cite{bib:uai,bib:unifai,bib:ropad,bib:advforget}. The models are trained in an alternating manner such that the discriminators are frozen when the rest of the model is updated and vice versa. While updating the discriminators, we use the standard categorical cross-entropy loss with object-types used as prediction targets. On the other hand, while training the rest of the model, we minimize (a) the cross-entropy loss for object-or-not classification, (b) smooth L\textsubscript{1} loss for bounding box regression, and (c) the negative entropy of discriminator predictions. This entropy maximization forces the upstream part of detection models to exclude object-type information from the features it outputs. The discriminator is updated five times for every update to the rest of the model and the negative entropy is weighted with a multiplier $\alpha$ (tuned on $\{0.1, 1\}$) in the overall objective. \Cref{fig:model} summarizes the complete framework.

During test-time inference, the discriminators are detached from the model, giving back the original network with the standard layers and parameter-count. Thus, our framework does not cause performance delays.

\subsection{Model Instantiation}

We demonstrate the applicability of the proposed adversarial framework to both Faster R-CNN (FRCNN), a two-stage detector, and SSD, a one-stage detector. We use the publicly available MMDetection~\cite{bib:mmdetection} framework for implementing the models and running experiments. We train two versions of the adversarial models -- one is trained from scratch and the other is finetuned from a pretrained baseline class-aware model. Further details are as follows.

\vspace{3pt}
\noindent\textbf{Faster R-CNN.}\quad The FRCNN model first generates regions of interest in the input image, which is followed by extracting features from and making object-type and bounding box predictions for each region. We create a class-agnostic adversarial version of FRCNN by (1) replacing the multi-class object-type classification layer with a binary object-or-not layer, and (2) attaching an adversarial discriminator on top of the feature extraction layer that provides inputs to the classification and regression heads. Thus, during training, this feature layer serves three prediction heads instead of the standard two. In our experiments, we use the standard FRCNN model available in MMDetection, which includes a ResNet-50 backbone and a Feature Pyramid Network~\cite{bib:fpn}.

\vspace{3pt}
\noindent\textbf{SSD.}\quad An SSD model utilizes features from several layers of its backbone network to detect objects of different scales, corresponding to the depth-levels of the backbone layers. Specifically, SSD models contain classification and regression layers for making predictions at each depth-level. In order to create a class-agnostic adversarial version of the SSD model, we (1) replace each object-type classification layer with a binary object-or-not layer, and (2) attach an adversarial discriminator at each depth-level where predictions are made. Thus, during training, each prediction level in the resulting model has three prediction heads instead of the conventional two. We use the standard SSD-300 with VGG-16 pretrained on ImageNet as the backbone.

{
\def \fs {0.33} % {0.44}
\def \sfs {1} % {0.66}
\begin{figure*}
\centering
% \captionsetup[figure]{aboveskip=-15pt}
\begin{subfigure}{\fs\textwidth}
\centering
\includegraphics[width=\sfs\textwidth]{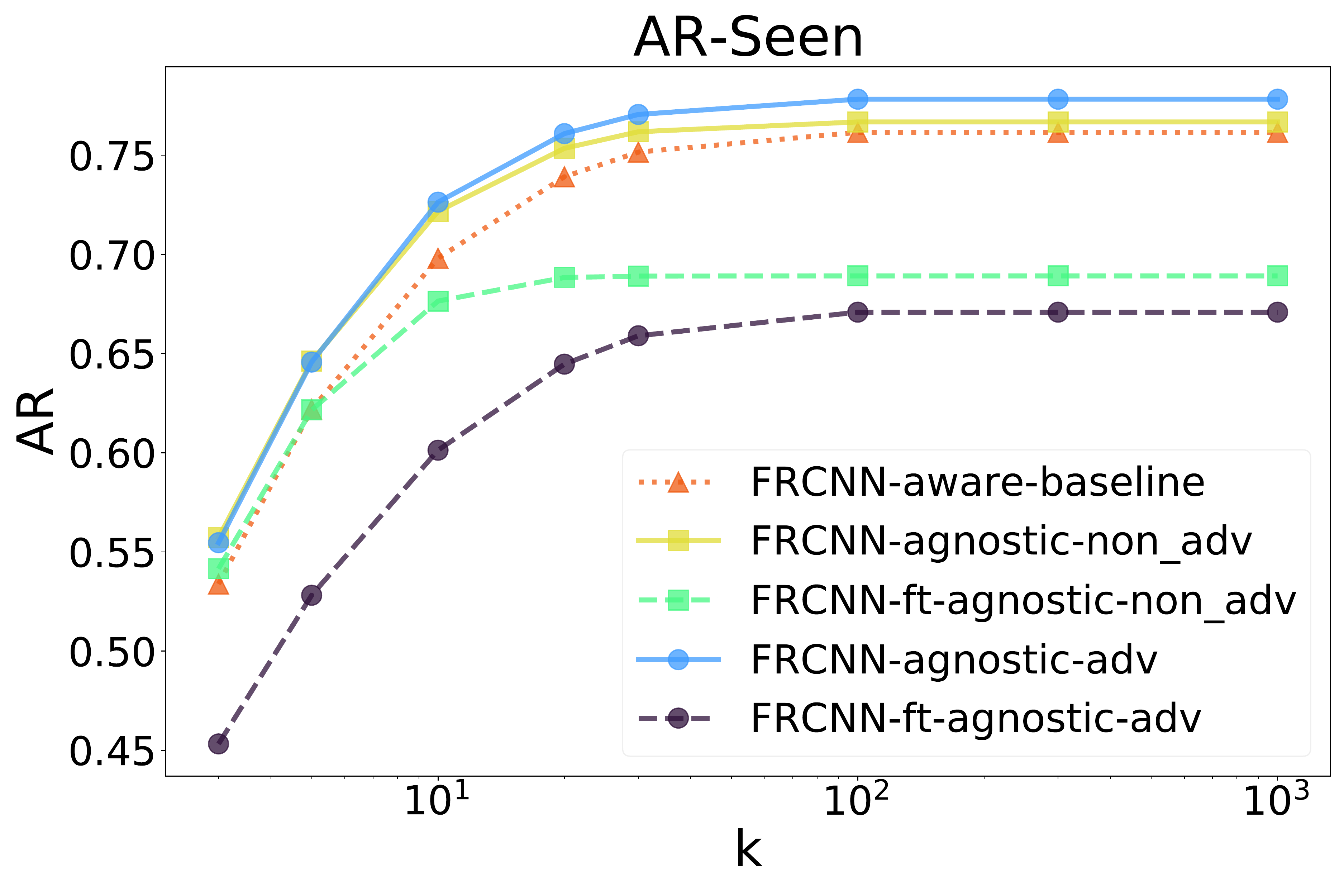}
% \caption{Conventional class-aware object detection}
\end{subfigure}
\begin{subfigure}{\fs\textwidth}
\centering
\includegraphics[width=\sfs\textwidth]{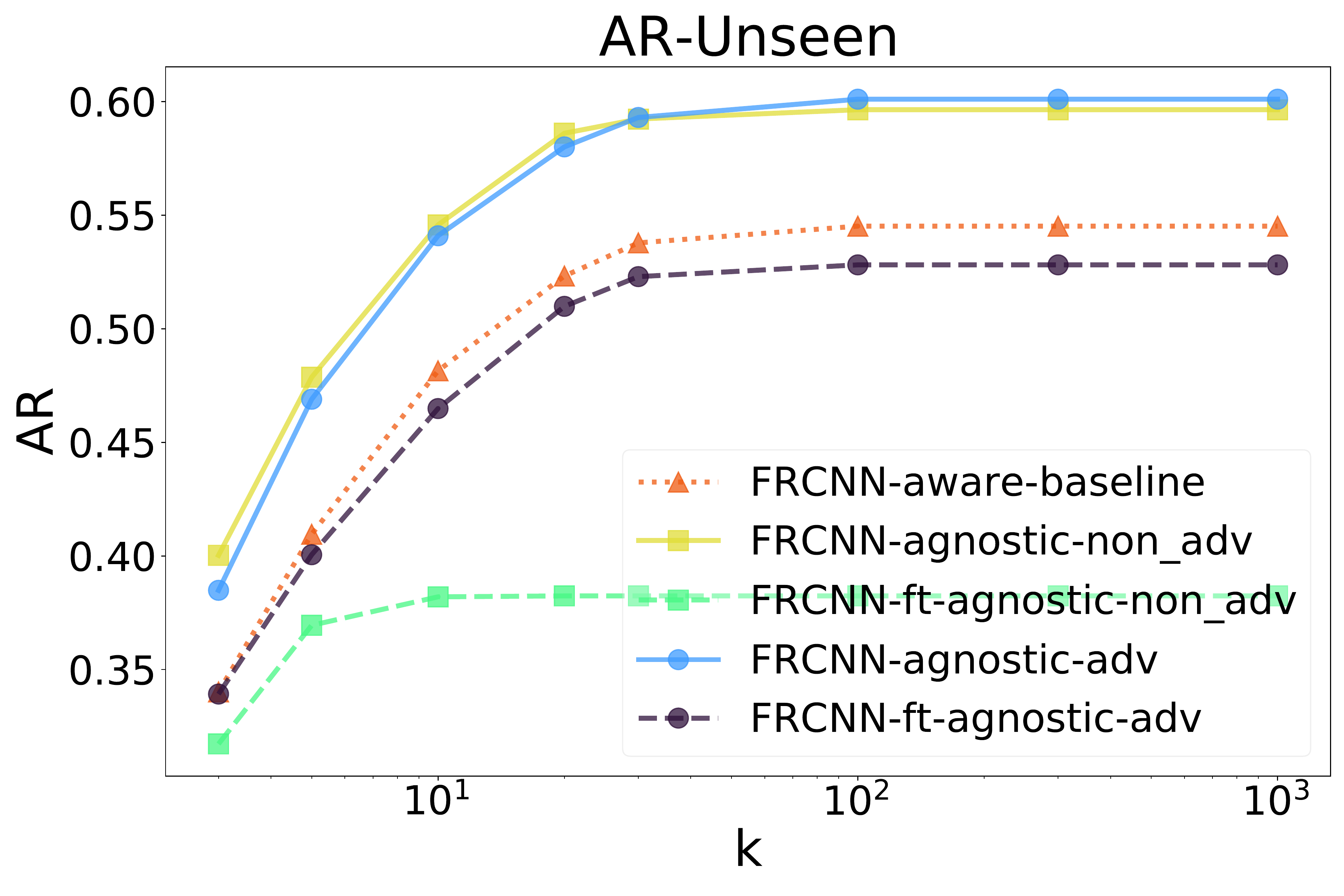}
% \caption{Proposed class-agnostic object detection}
\end{subfigure}
\begin{subfigure}{\fs\textwidth}
\centering
\includegraphics[width=\sfs\textwidth]{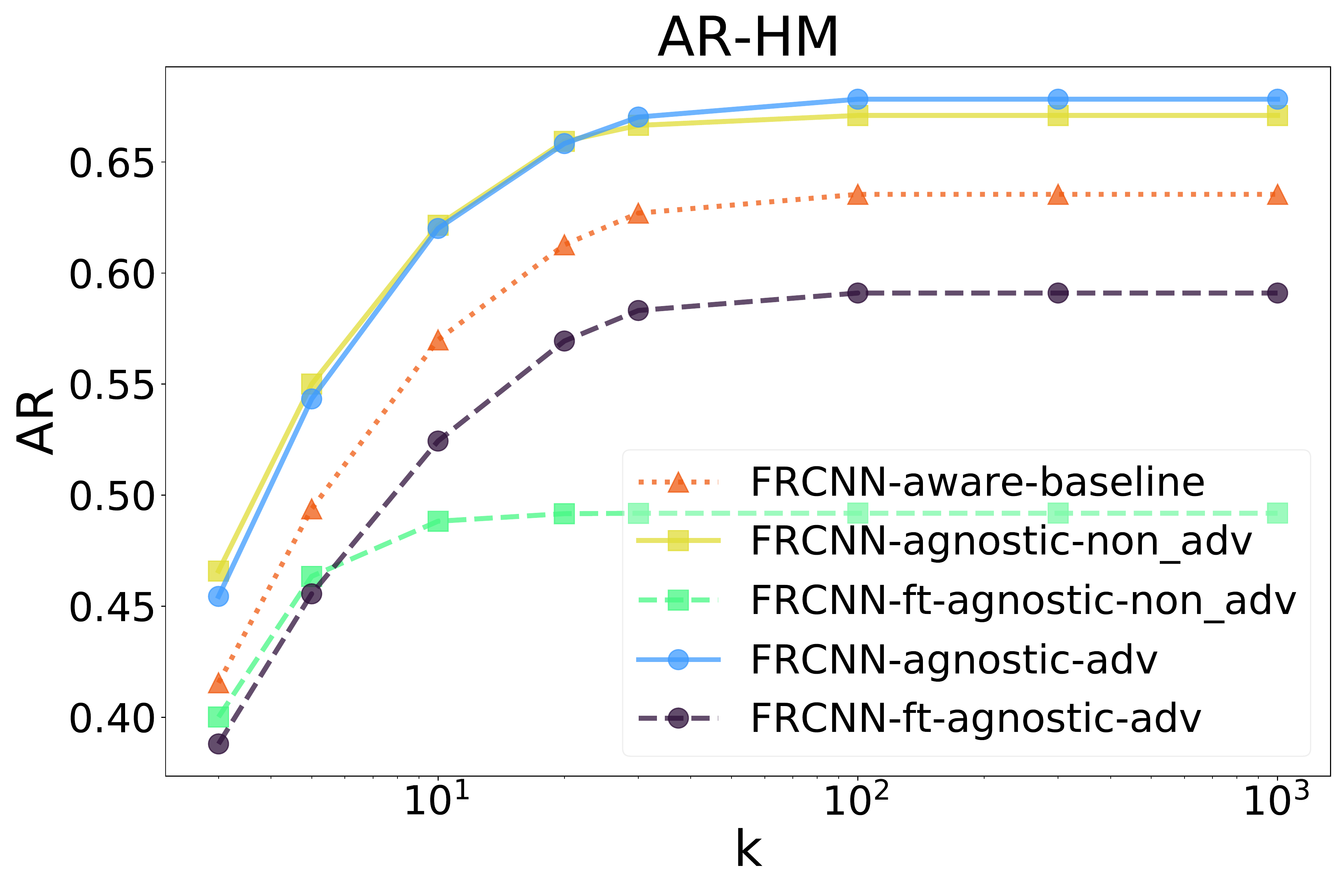}
% \caption{Proposed class-agnostic object detection}
\end{subfigure}
\begin{subfigure}{\fs\textwidth}
\centering
\includegraphics[width=\sfs\textwidth]{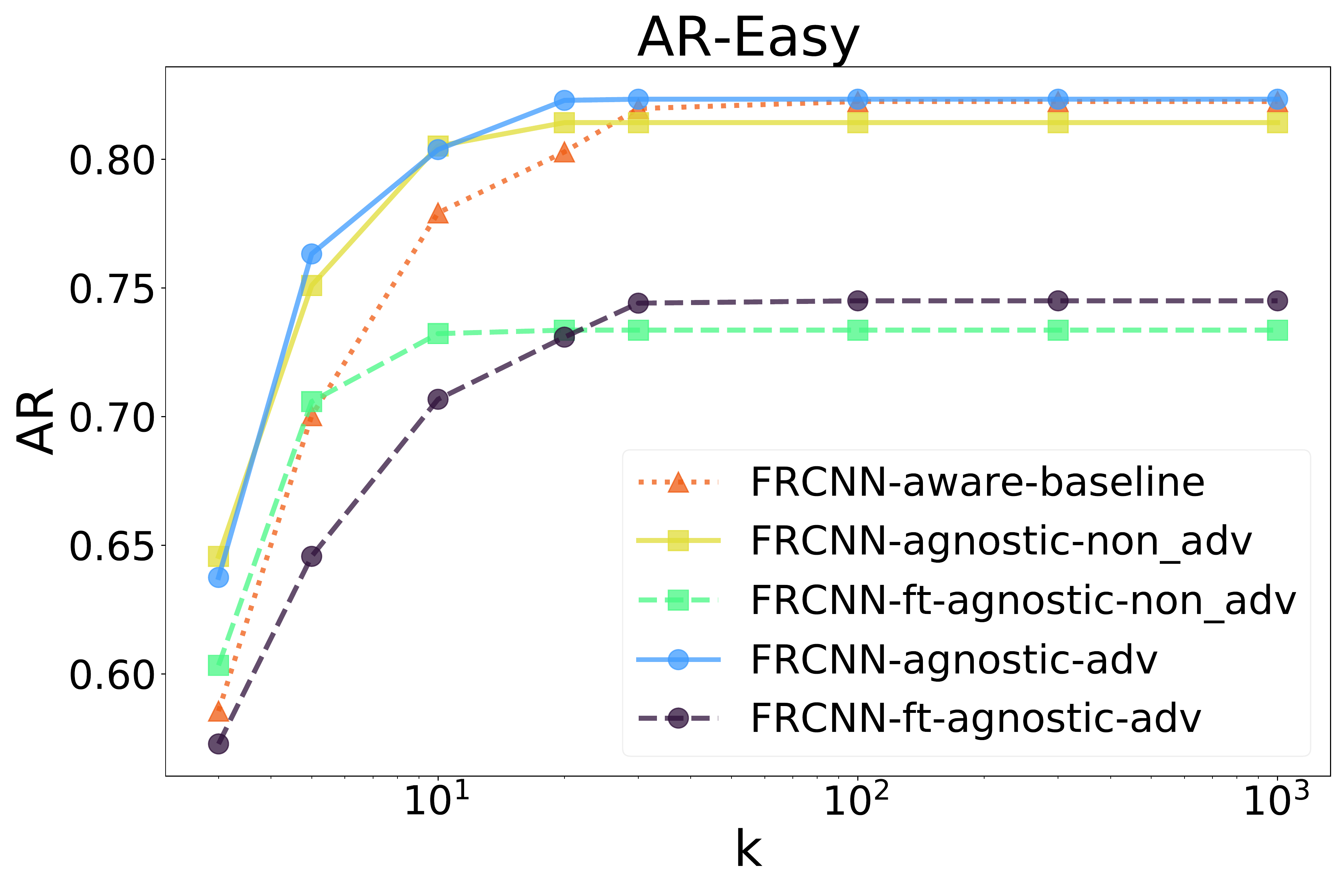}
% \caption{Conventional class-aware object detection}
\end{subfigure}
\begin{subfigure}{\fs\textwidth}
\centering
\includegraphics[width=\sfs\textwidth]{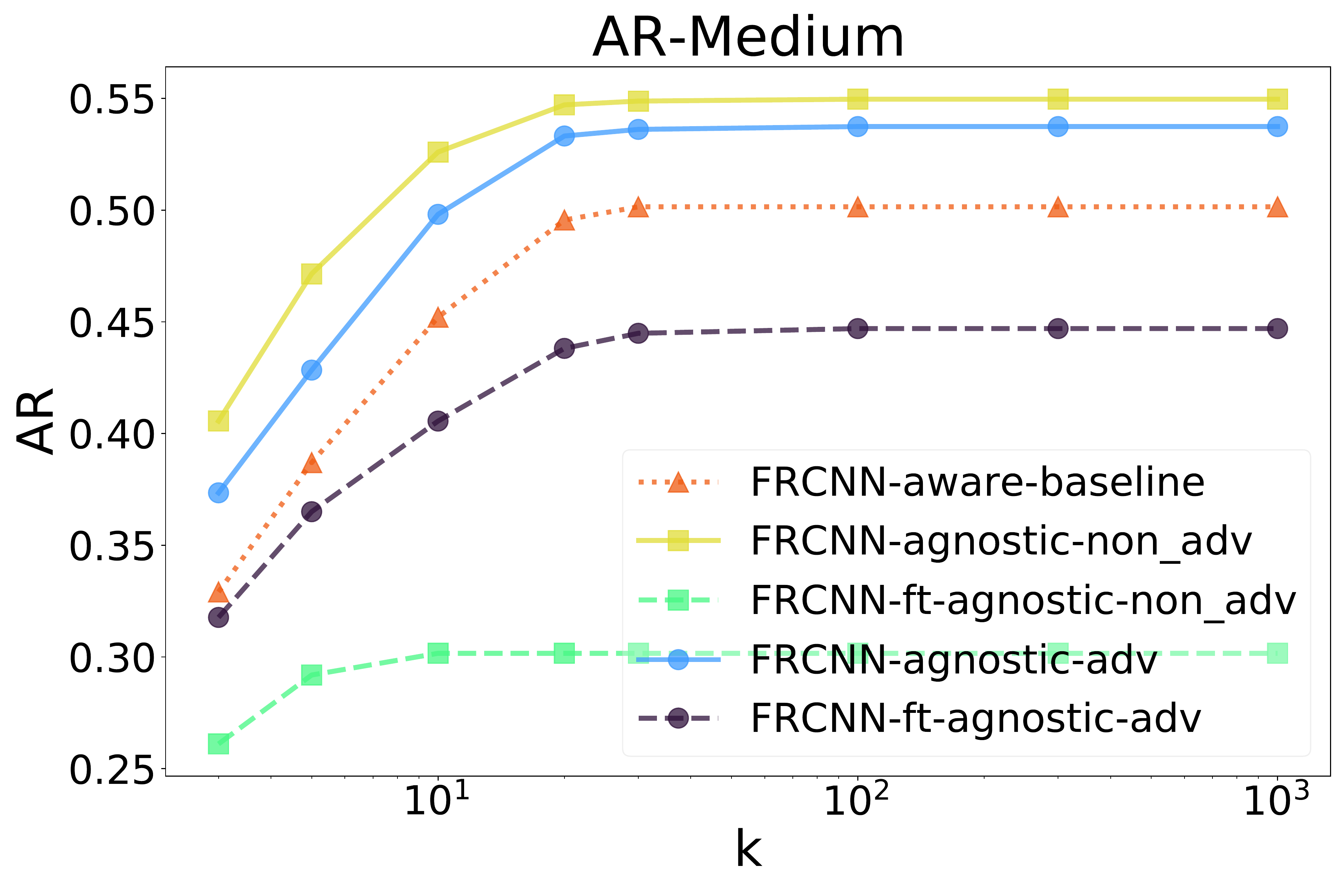}
% \caption{Proposed class-agnostic object detection}
\end{subfigure}
\begin{subfigure}{\fs\textwidth}
\centering
\includegraphics[width=\sfs\textwidth]{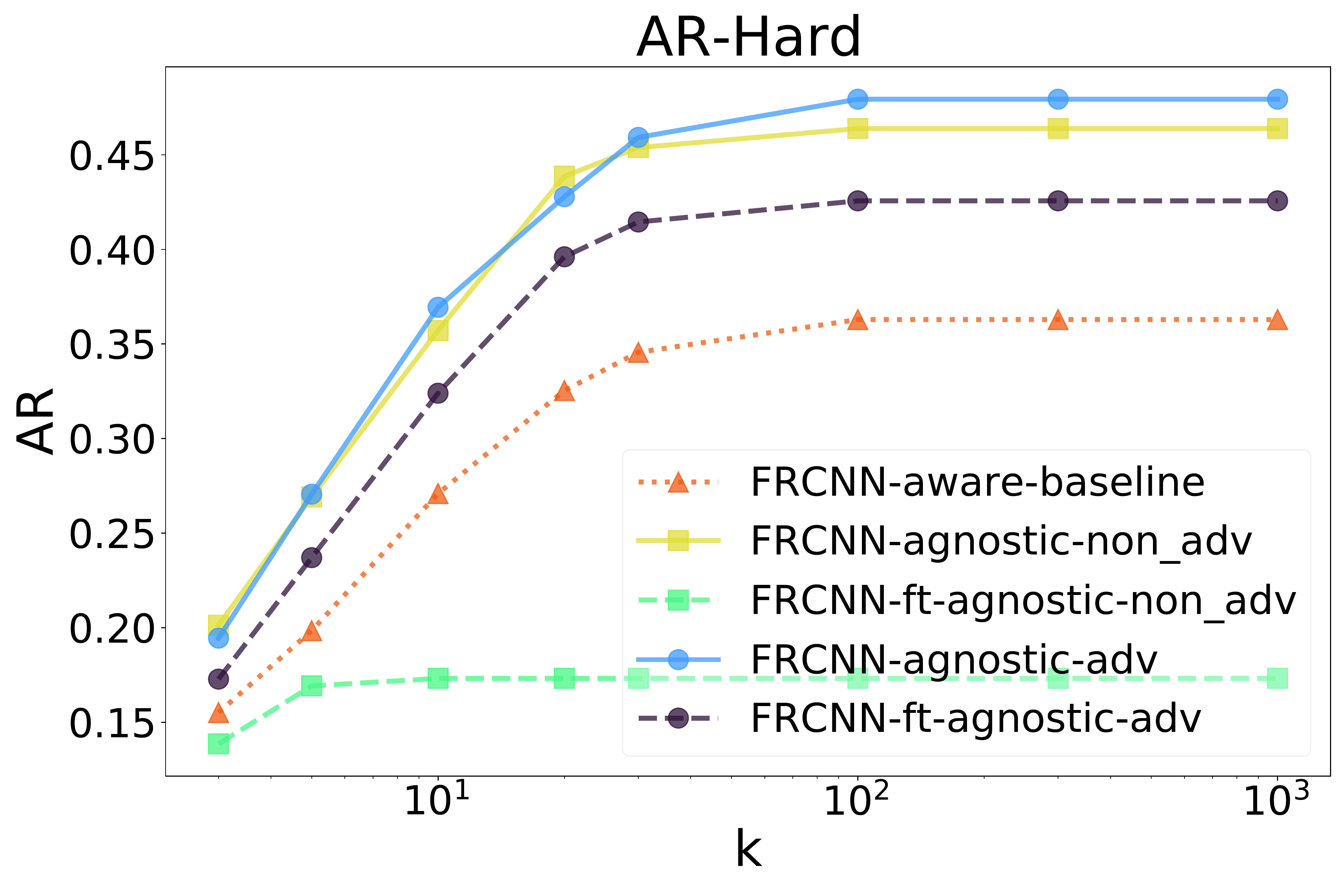}
% \caption{Proposed class-agnostic object detection}
\end{subfigure}
\begin{subfigure}{\fs\textwidth}
\centering
\includegraphics[width=\sfs\textwidth]{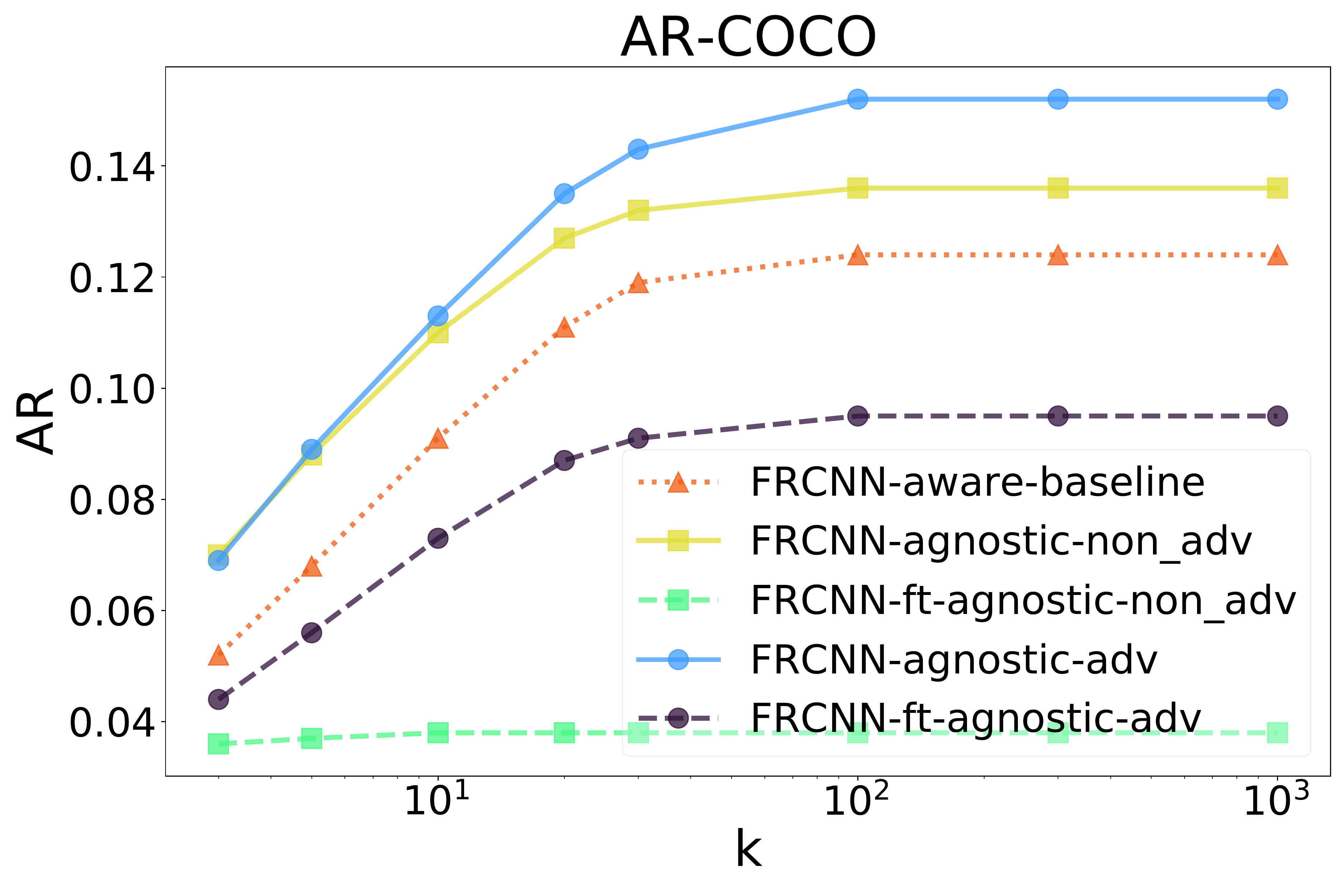}
% \caption{Conventional class-aware object detection}
\end{subfigure}
\begin{subfigure}{\fs\textwidth}
\centering
\includegraphics[width=\sfs\textwidth]{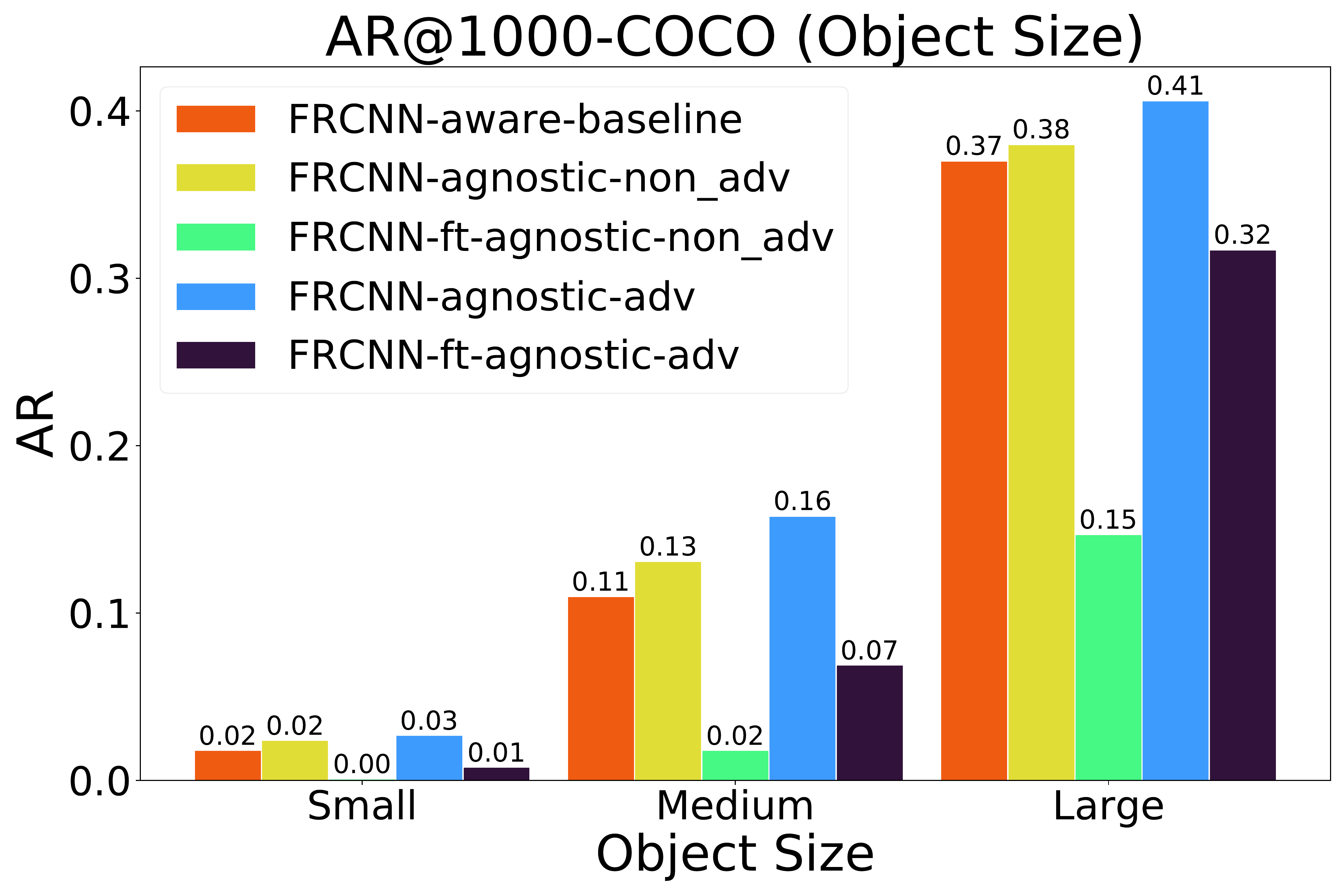}
% \caption{Proposed class-agnostic object detection}
\end{subfigure}
\begin{subfigure}{\fs\textwidth}
\scriptsize
\centering
\vspace{-20pt}
\caption*{Terminology}
\setlength{\tabcolsep}{0.5em}
\begin{tabular}{c c}
\toprule
    \textbf{Token} & \textbf{Meaning} \\
    \cmidrule(r){1-1} \cmidrule(l){2-2}
    -aware- & Class-aware model \\
    -agnostic- & Class-agnostic model \\
    -ft- & Finetuned from FRCNN-aware-baseline \\
    -non\_adv & Non-adversarial model \\
    -adv & Adversarially trained model \\
\bottomrule
\end{tabular}
\end{subfigure}
\caption{Generalization results for FRCNN models trained on the seen VOC dataset. The top row shows macro-level AR@$k$ for seen and unseen classes in VOC and their harmonic mean (AR-HM). FRCNN-agnostic-adv performs the best overall. The second row shows micro-level results for the easy, medium, and hard unseen classes. FRCNN-agnostic-adv performs the best on the hard and easy classes with recall drop for the medium class. The last row provides results of evaluation on the COCO data of 60 unseen classes. FRCNN-agnostic-adv achieves the best AR@$k$ for objects of all sizes.}
\label{fig:frcnn_voc}
\vspace{-7pt}
\end{figure*}
}

%% file: sections/05_evaluation.tex
\section{Training and Evaluation Protocols}
\label{sec:protocols}

We propose two kinds of experiments for evaluating class-agnostic detection, geared towards measuring (1) generalization to unseen object types, and (2) the downstream utility of trained models. While there are several ways to design experiments for (1) and (2), we propose first steps with potential for refinement in future work.

\subsection{Generalization to Unseen Object-types.}
\label{subsec:gen}

Class-agnostic detectors should, by definition, not be limited to object types seen during training. Hence, it is important to evaluate their efficacy at identifying unseen types of objects. We measure this performance as average recall (AR; also known as detection rate~\cite{bib:edge_proposal_3_edgeboxes}) at various number ($k \in \{3, 5, 10, 20, 30, 100, 300, 1000\}$) of allowed bounding box predictions, i.e., AR@$k$ ($\text{IoU} \ge 0.5$). The goal here is to achieve high AR@$k$ at all $k$ levels. We design two sets of generalization experiments based on this setup.

\vspace{3pt}
\noindent\textbf{Experiment I.}\quad We split the VOC dataset into 17 seen classes and three unseen classes. Seen classes in the VOC 07+12 training set are used for learning the models while both seen and unseen classes in the VOC 07 validation set are used for evaluation as AR@$k$ of seen and unseen classes. The harmonic mean of seen and unseen AR@$k$, for each $k$, is reported as a measure of overall performance.

In order to select the three unseen classes, we compute the confusion matrix\footnote{https://github.com/kaanakan/object\_detection\_confusion\_matrix} of a 300-layer SSD model trained on the standard 20-class VOC 07+12 dataset, and use F\textsubscript{1} scores to determine easy, medium, and hard classes. Specifically, the class with the highest F\textsubscript{1} is the hardest for generalization as it is the most distinguishable from the other classes, making it the most difficult to generalize to if it were not seen during training. Similarly, the class with the lowest F\textsubscript{1} is picked as the easy class. The medium class is the one with the median F\textsubscript{1}. Thus, the unseen set comprises \textit{cow} (easy), \textit{boat} (medium) and \textit{tvmonitor} (hard) classes, while the rest are considered seen. In addition to the macro AR@$k$ scores, micro AR@$k$ are reported for each unseen class. Besides the evaluation on VOC, the (VOC) models are tested for generalization on the 60 non-VOC classes in the COCO 2017 validation set using the same AR@$k$ metric. AR@1000 is further reported for small, medium, and large sized objects belonging to the 60 classes.

\begin{table*}
\centering
\caption{\label{tab:generalization}Generalization results for experiments I and II. AR@1000-Unseen shows results for unseen VOC classes, AR@1000-COCO for unseen COCO classes, and AR@1000-OI for unseen Open Images classes. ``Ovr'', ``Med'', ``Sml'', and ``Lrg'' stand for overall, medium, small, and large, respectively. ``-aw-'' and ``-ag-'' in the model name indicate whether the model is class-aware or -agnostic, ``-ft-'' tells whether the model was finetuned from the class-aware baseline, and ``-ad'' represents models trained adversarially. FRCNN-aw-prop refers to the evaluation of the FRCNN-aw proposals from its first stage Region Proposal Network. Class-agnostic models generalize better than class-aware models with those trained adversarially from scratch performing the best overall.}
\setlength{\tabcolsep}{0.55em} % for horizontal padding
\begin{tabular}{l c c c c c c c c c c c c}
 \toprule
  & \multicolumn{8}{c}{\textbf{I: Training on seen VOC}} & \multicolumn{4}{c}{\textbf{II: Training on COCO}} \\
  \cmidrule(lr){2-9} \cmidrule(lr){10-13}
  & \multicolumn{4}{c}{\textbf{AR@1000-Unseen}} & \multicolumn{4}{c}{\textbf{AR@1000-COCO}} & \multicolumn{4}{c}{\textbf{AR@1000-OI}} \\
 \cmidrule(lr){2-5} \cmidrule(lr){6-9} \cmidrule(lr){10-13}
\textbf{Model} & Ovr & Easy & Med & Hard & Ovr & Sml & Med & Lrg & Ovr & Sml & Med & Lrg \\
\cmidrule(r){1-1} \cmidrule(lr){2-2} \cmidrule(lr){3-3} \cmidrule(lr){4-4} \cmidrule(lr){5-5} \cmidrule(lr){6-6} \cmidrule(lr){7-7} \cmidrule(lr){8-8}  \cmidrule(lr){9-9}  \cmidrule(lr){10-10} \cmidrule(lr){11-11} \cmidrule(lr){12-12} \cmidrule(lr){13-13}
 FRCNN-aw-prop & 0.374 & 0.388 & 0.383 & 0.355 & 0.107 & 0.006 & 0.071 & 0.375 & 0.153 & 0.005 & 0.037 & 0.329 \\
 FRCNN-aw & 0.545 & 0.822 & 0.501 & 0.363 & 0.124 & 0.018 & 0.110 & 0.370 & 0.170 & 0.016 & 0.083 & 0.330 \\
 FRCNN-ft-ag & 0.382 & 0.733 & 0.302 & 0.173 & 0.038 & 0.001 & 0.018 & 0.147 & 0.184 & \textbf{0.028} & \textbf{0.138} & 0.313 \\
 FRCNN-ag & 0.596 & 0.814 & \textbf{0.550} & 0.464 & 0.136 & 0.024 & 0.131 & 0.380 & 0.182 & \textbf{0.028} & 0.125 & 0.318 \\
 FRCNN-ft-ag-ad & 0.528 & 0.745 & 0.447 & 0.426 & 0.095 & 0.008 & 0.069 & 0.317 & 0.181 & 0.026 & 0.112 & 0.328 \\
 FRCNN-ag-ad & \textbf{0.601} & \textbf{0.823} & 0.537 & \textbf{0.479} & \textbf{0.152} & \textbf{0.027} & \textbf{0.158} & \textbf{0.406} & \textbf{0.192} & \textbf{0.028} & 0.131 & \textbf{0.337} \\
 \cmidrule(r){1-1} \cmidrule(lr){2-9} \cmidrule(lr){10-13}
 SSD-aw & 0.663 & 0.831 & 0.594 & 0.588 & 0.156 & 0.019 & 0.194 & 0.391 & 0.199 & 0.013 & 0.145 & \textbf{0.350} \\
 SSD-ft-ag & 0.671 & 0.819 & 0.668 & 0.555 & 0.177 & \textbf{0.040} & 0.238 & 0.366 & 0.188 & 0.020 & 0.168 & 0.301 \\
 SSD-ag & 0.702 & 0.810 & 0.669 & 0.645 & 0.181 & 0.031 & 0.239 & 0.404 & 0.206 & 0.023 & 0.206 & 0.314 \\
 SSD-ft-ag-ad & 0.683 & 0.809 & 0.651 & 0.612 & 0.155 & 0.028 & 0.197 & 0.354 & 0.194 & 0.018 & 0.181 & 0.308 \\
 SSD-ag-ad & \textbf{0.722} & \textbf{0.833} & \textbf{0.689} & \textbf{0.660} & \textbf{0.196} & 0.036 & \textbf{0.252} & \textbf{0.443} & \textbf{0.210} & \textbf{0.024} & \textbf{0.211} & 0.319 \\
\bottomrule
\end{tabular}
\vspace{-5pt}
\end{table*}

\vspace{1pt}
\noindent\textbf{Experiment II.}\quad In this experiment, models are trained on the COCO 2017 training set and evaluated on non-overlapping classes in the Open Images V6 test set. We first normalize the class names in both COCO and Open Images to find exact matches between classes in the two datasets. We identify a few classes that have slightly different names in the two datasets and normalize them manually. We then build a semantic tree using the publicly available Open Images class hierarchy and exclude all classes that either match exactly with a COCO class or have a COCO class as a predecessor or successor. The remaining classes are used for testing. AR@$k$ is used as the evaluation metric.

\subsection{Downstream Utility}
\label{subsec:downstream}

A major motivation for research in the domain of class-agnostic object detection is the potential for widespread downstream utilization of class agnostic bounding boxes for extracting objects in images and using them for various visual reasoning tasks. In this work, we propose evaluation of downtream utility in terms of object recognition on images in the ObjectNet~\cite{bib:objectnet} dataset. Detection models are used to predict $M$ bounding boxes for each image, which are then use to crop the image into $M$ versions. Pretrained\footnote{https://pytorch.org/docs/stable/torchvision/models.html} ImageNet classifiers are then used to predict the object class from the cropped images. Here we use ResNet-152~\cite{bib:resnet}, MobileNet-v2~\cite{bib:mobilenetv2}, and Inception-v3~\cite{bib:inceptionv3} models.

Two metrics are used for evaluation -- Accuracy@$M$ for $M \in \{1, 5, 10\}$, and Best-overlap (BO) accuracy. For Accuracy@$M$, the classifier's prediction on at least one of the $M$ crops needs to be correct for the image to be considered as successfully classified. BO-accuracy is calculated using bounding boxes that have the highest intersection over union (IoU) with the ground-truth~\cite{bib:objectnet_reanalysis} bounding boxes.

{
\def \fs {0.33} % {0.44}
\def \sfs {1} % {0.66}
\begin{figure*}
\centering
% \captionsetup[figure]{aboveskip=-20pt}
\begin{subfigure}{\fs\textwidth}
\centering
\includegraphics[width=\sfs\textwidth]{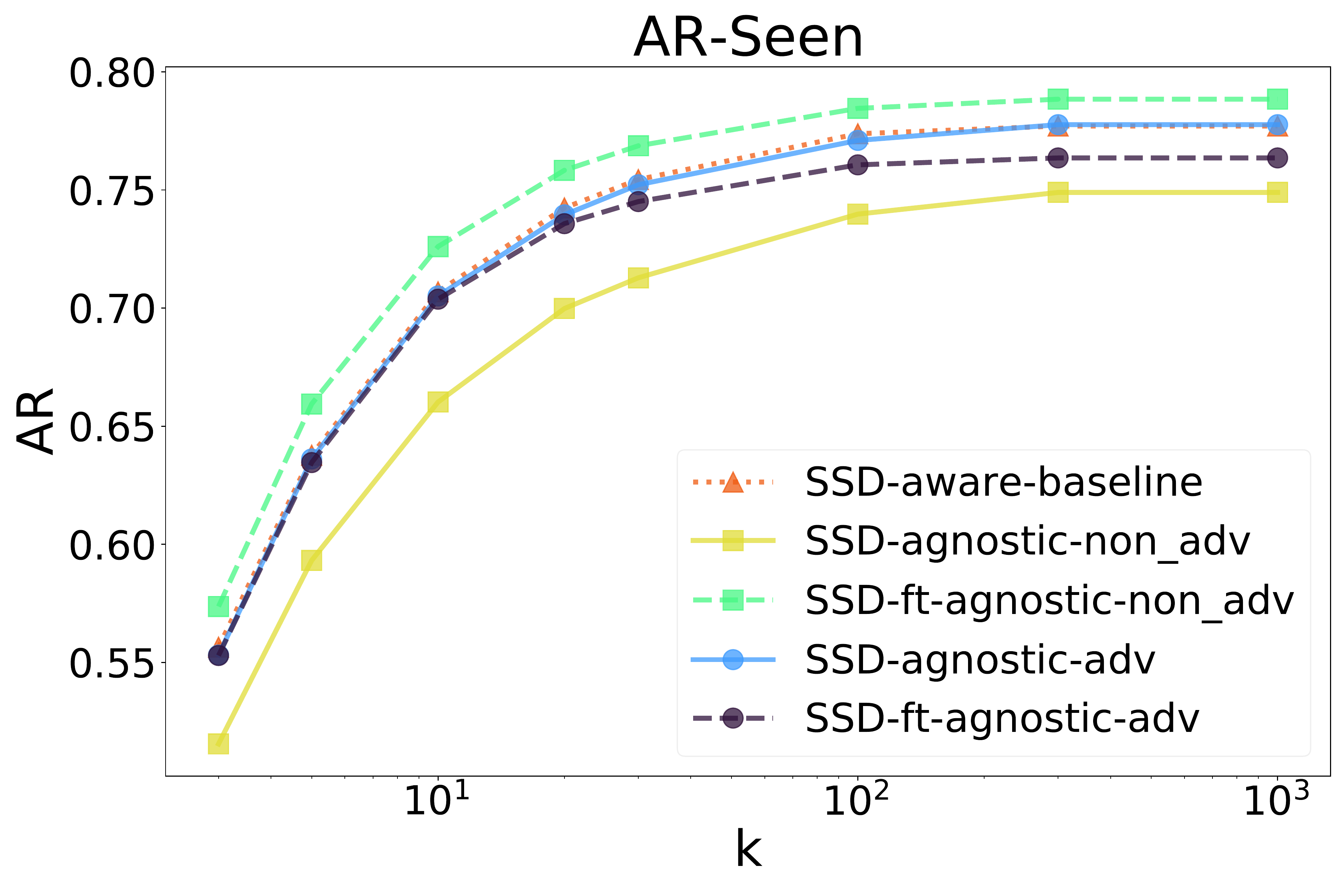}
% \caption{Conventional class-aware object detection}
\end{subfigure}
\begin{subfigure}{\fs\textwidth}
\centering
\includegraphics[width=\sfs\textwidth]{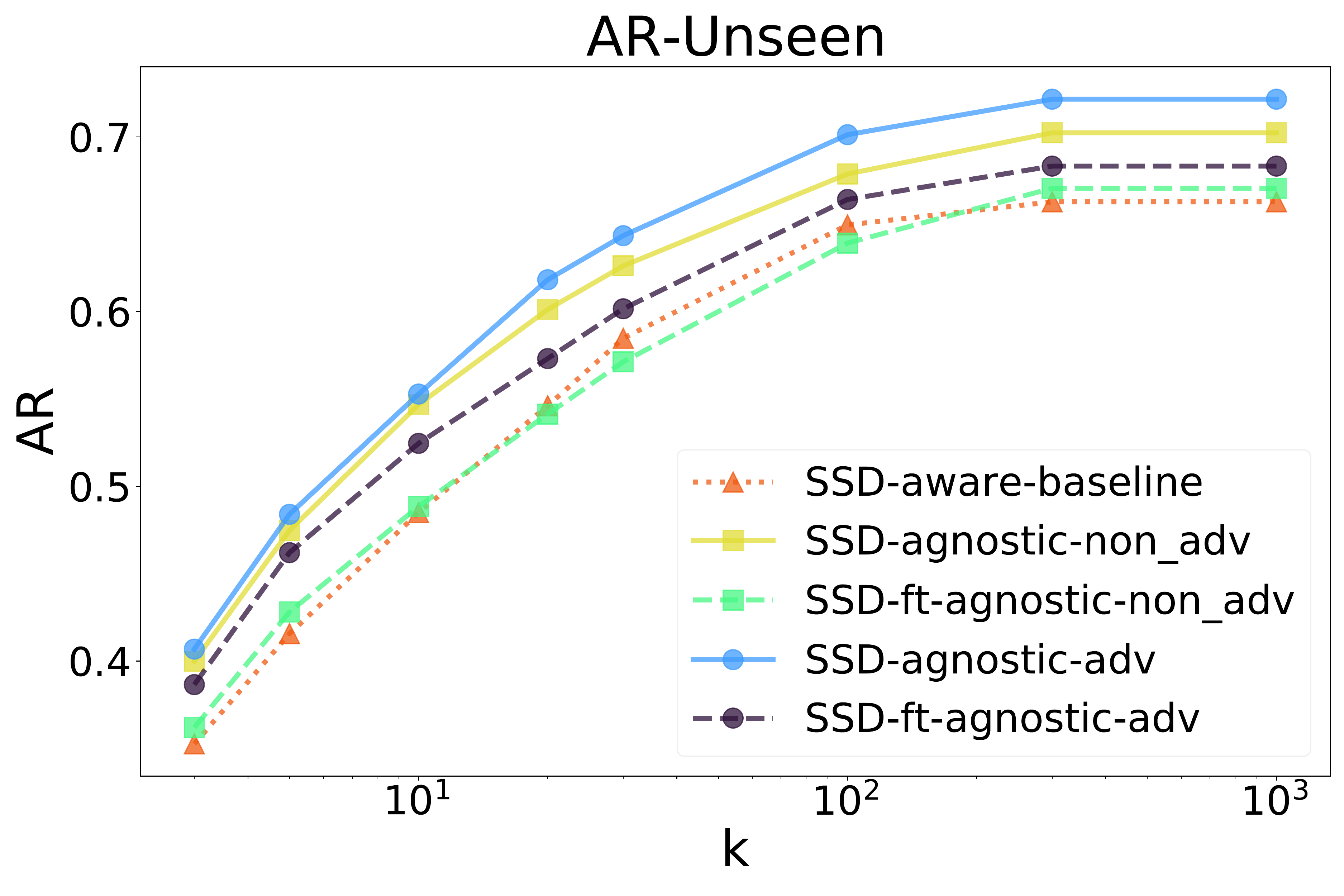}
% \caption{Proposed class-agnostic object detection}
\end{subfigure}
\begin{subfigure}{\fs\textwidth}
\centering
\includegraphics[width=\sfs\textwidth]{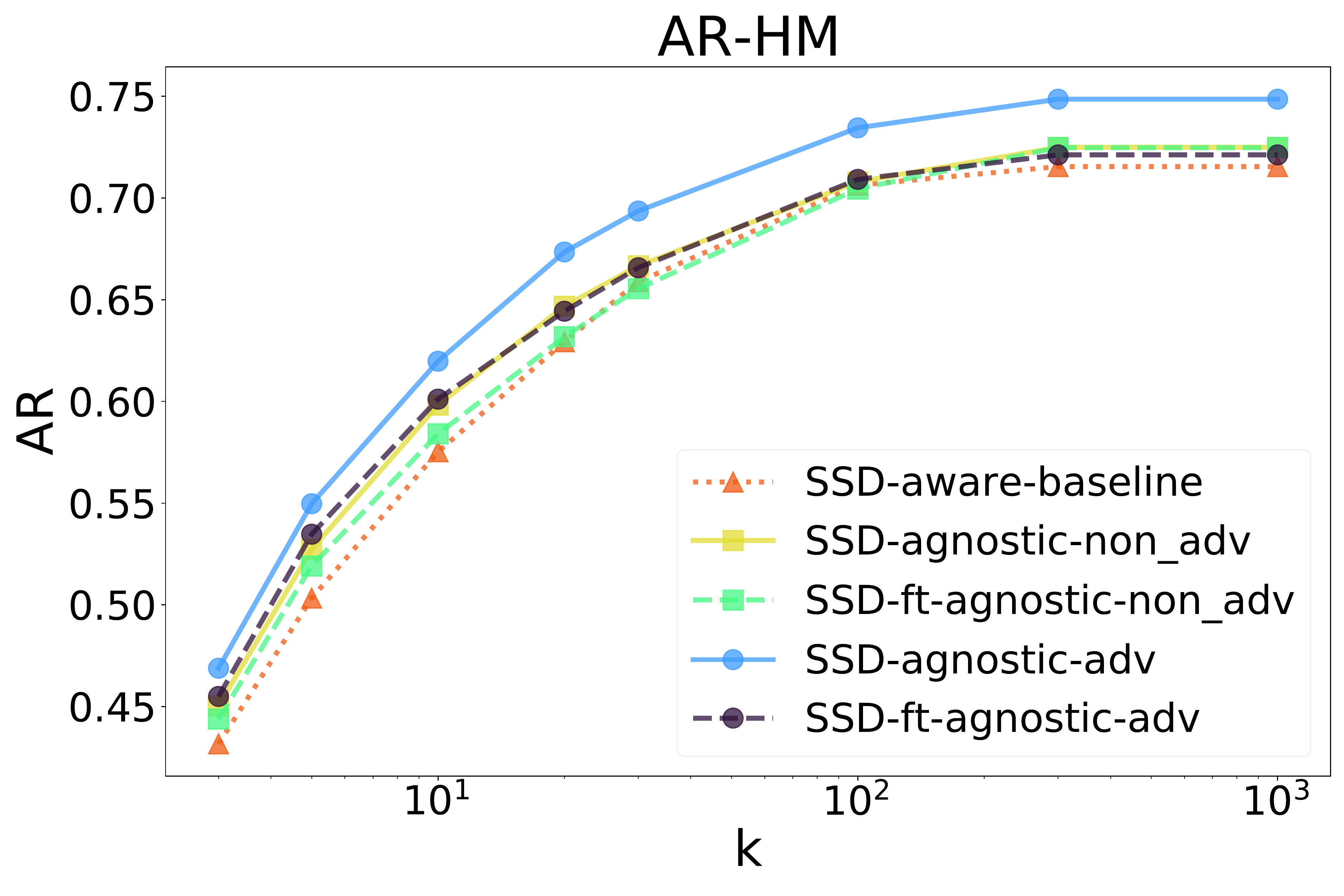}
% \caption{Proposed class-agnostic object detection}
\end{subfigure}
\begin{subfigure}{\fs\textwidth}
\centering
\includegraphics[width=\sfs\textwidth]{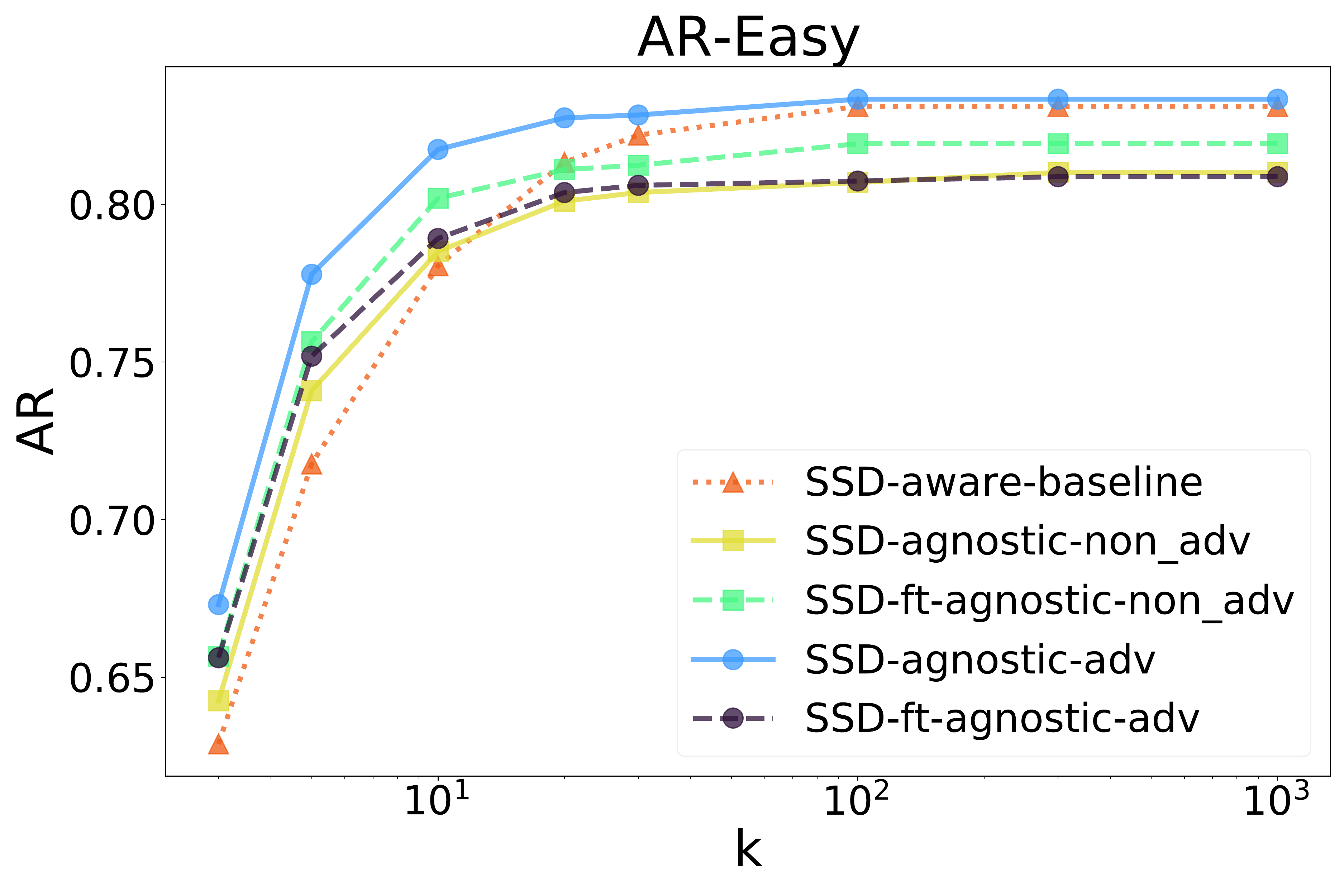}
% \caption{Conventional class-aware object detection}
\end{subfigure}
\begin{subfigure}{\fs\textwidth}
\centering
\includegraphics[width=\sfs\textwidth]{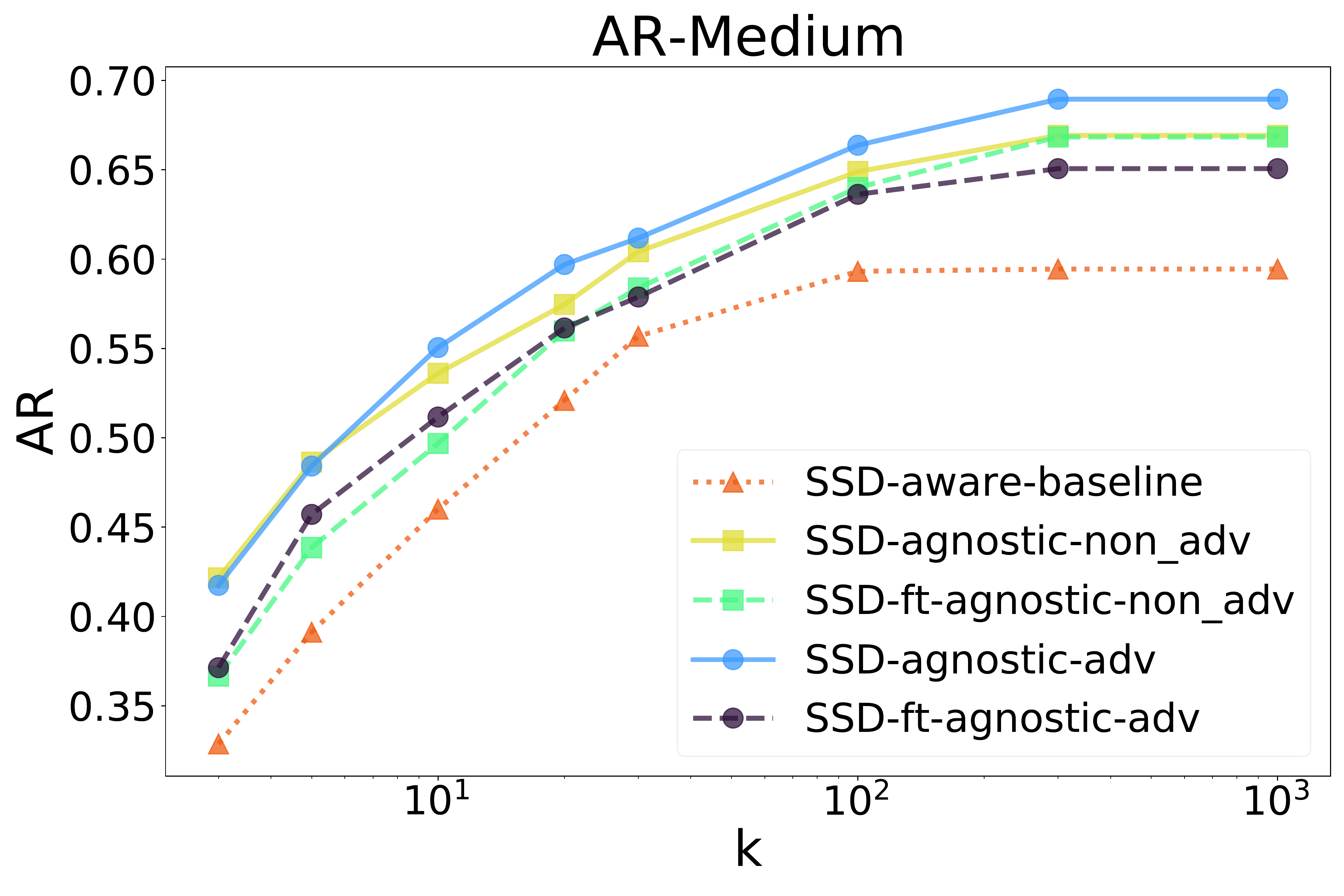}
% \caption{Proposed class-agnostic object detection}
\end{subfigure}
\begin{subfigure}{\fs\textwidth}
\centering
\includegraphics[width=\sfs\textwidth]{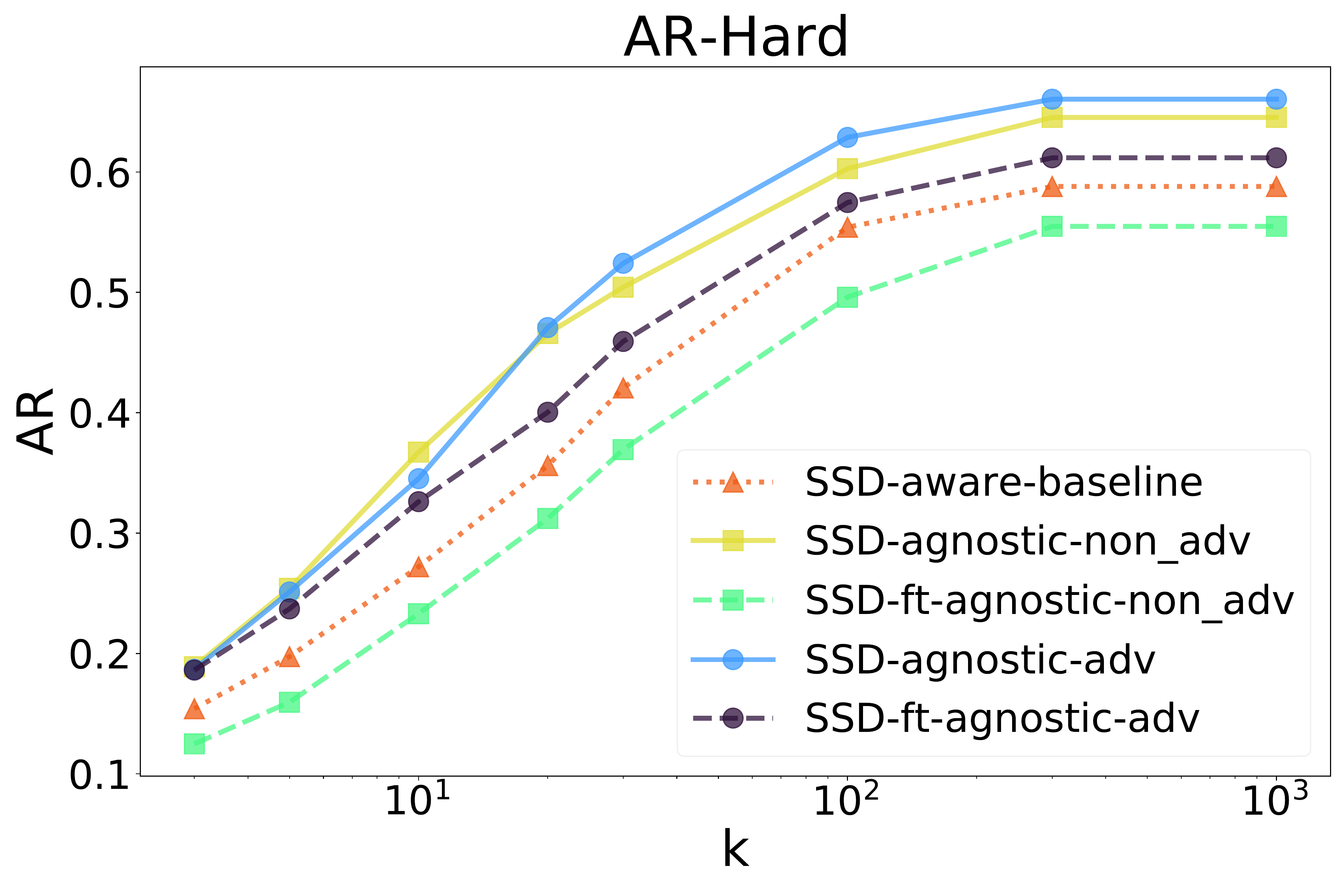}
% \caption{Proposed class-agnostic object detection}
\end{subfigure}
\begin{subfigure}{\fs\textwidth}
\centering
\includegraphics[width=\sfs\textwidth]{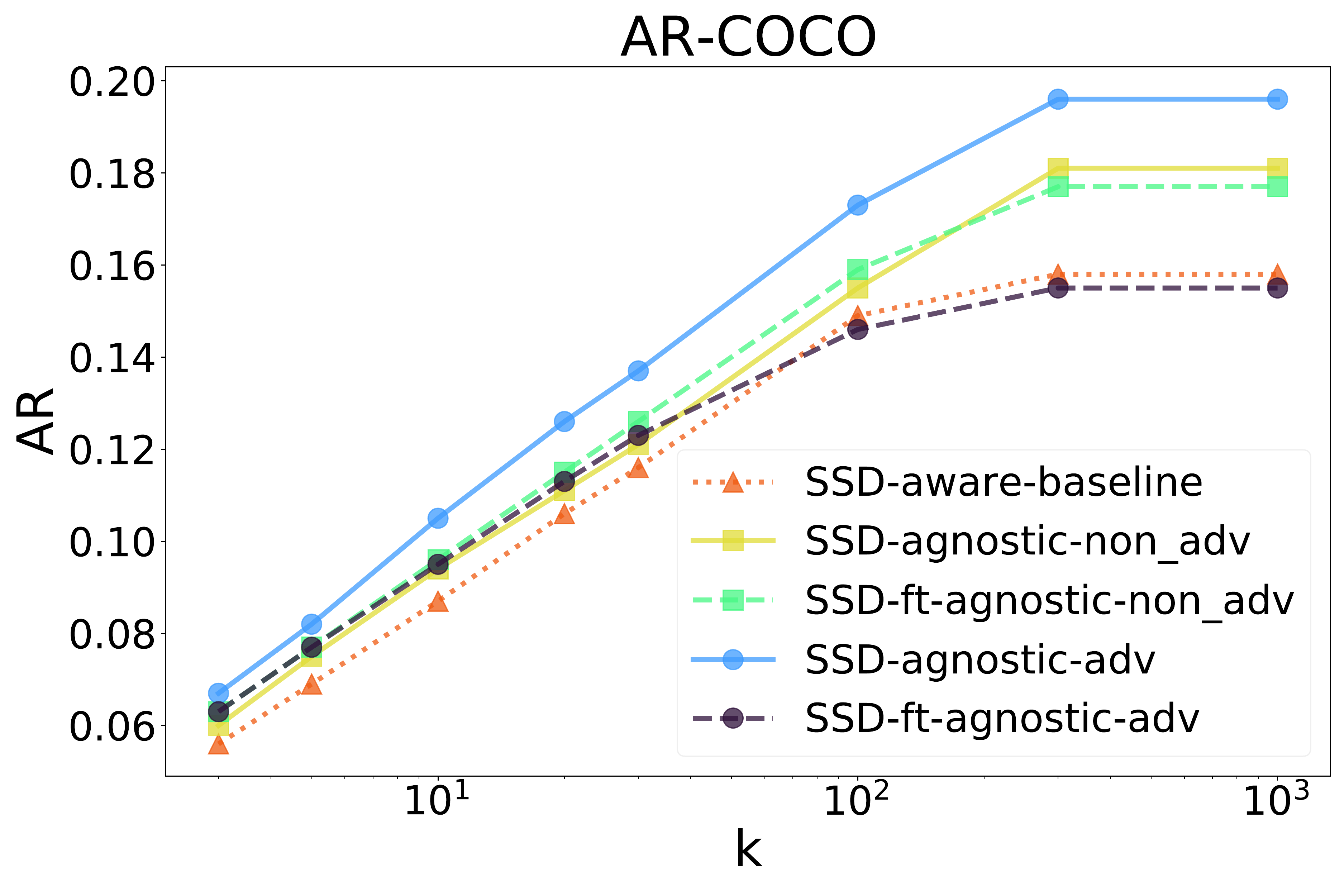}
% \caption{Conventional class-aware object detection}
\end{subfigure}
\begin{subfigure}{\fs\textwidth}
\centering
\includegraphics[width=\sfs\textwidth]{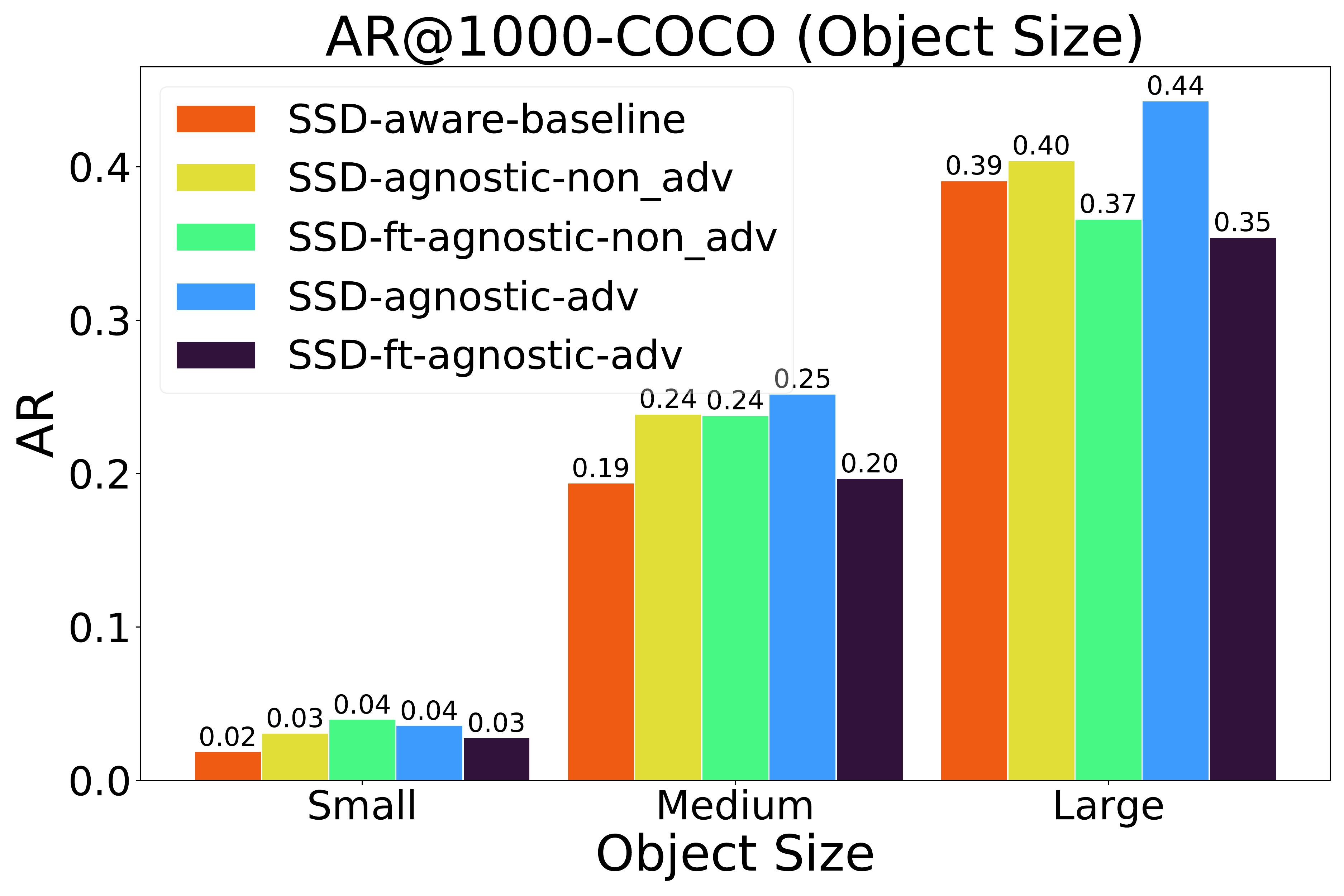}
% \caption{Proposed class-agnostic object detection}
\end{subfigure}
\begin{subfigure}{\fs\textwidth}
\scriptsize
\centering
\vspace{-20pt}
\caption*{Terminology}
\setlength{\tabcolsep}{0.7em}
\begin{tabular}{c c}
\toprule
    \textbf{Token} & \textbf{Meaning} \\
    \cmidrule(r){1-1} \cmidrule(l){2-2}
    -aware- & Class-aware model \\
    -agnostic- & Class-agnostic model \\
    -ft- & Finetuned from SSD-aware-baseline \\
    -non\_adv & Non-adversarial model \\
    -adv & Adversarially trained model \\
\bottomrule
\end{tabular}
\end{subfigure}
\caption{Generalization results for SSD models trained on the seen VOC dataset. The top row shows macro-level AR@$k$ for seen and unseen classes in VOC as well as their harmonic mean (AR-HM). SSD-agnostic-adv performs the best on AR-Unseen and AR-HM, with a drop in AR-Seen, but the models that outperform SSD-agnostic-adv on AR-Seen do significantly worse on AR-Unseen and AR-HM. The second row shows micro-level results for the easy, medium, and hard unseen classes. SSD-agnostic-adv performs the best in all categories. The last row provides results of evaluation on the COCO data of 60 unseen classes. SSD-agnostic-adv achieves the best AR@$k$ with a slight reduction for small-sized objects.}
\label{fig:ssd_voc}
\vspace{-5pt}
\end{figure*}
}

\section{Results}
\label{sec:evaluation}

\subsection{Generalization to Unseen Object-types}
\label{subsec:results_gen}

\noindent\textbf{Experiment I.}\quad We train the baseline models and the adversarial models on the VOC seen set and evaluate them on both the VOC set and the COCO unseen classes. Results of this experiment with FRCNN as the base model are presented in \Cref{fig:frcnn_voc}. Class-agnostic models outperform standard class-aware models in overall performance, especially for unseen VOC classes. Breakdown of recall for unseen classes reveals that the margin in recall becomes larger as the difficulty of generalizability is increased, with the largest performance gains on the hard unseen class. Results on the COCO unseen classes show that the adversarial model performs the best overall and for objects of all sizes.

\Cref{fig:ssd_voc} summarizes the results with SSD as the base model. The results show that the class-agnostic SSD models largely outperform the conventional class-aware SSD model. Furthermore, the proposed adversarial model performs the best overall with some reduction in average recall for the VOC seen set. Breakdown of recall for the easy, medium, and hard unseen classes shows that the adversarial model performs the best across the board. Finally, results on COCO show consistent improvements from adversarial learning, especially for medium and large sized objects.

Hence, class-agnostic models perform better than class-aware models in generalizing to objects of unseen types and the proposed adversarial model performs the best overall.

\vspace{3pt}
\noindent\textbf{Experiment II.}\quad In this experiment, the baseline and the adversarial models are trained on the COCO dataset and evaluated on the non-overlapping classes of the Open Images, as outlined in \Cref{subsec:gen}. \Cref{fig:ssd_frcnn_coco} summarizes the results for both FRCNN and SSD models. In both cases, the class-agnostic models generalize better than the class-aware models overall, with the proposed class-agnostic models trained from scratch achieving the best AR scores.

\vspace{2pt}
\Cref{tab:generalization} presents AR@1000 generalization scores for experiments I and II, showing that the adversarially trained (from scratch) models perform the best in both settings.

% \vspace{3pt}
 We find that the recall of models that are finetuned from the pretrained class-aware baselines are worse than those that are trained from scratch. We attribute this to the difficulty of unlearning discriminative features for object-type classification and realigning to learn type-agnostic features, which prevents the finetuned models from achieving the same performance as those that are trained from scratch.

{
\def \fs {0.33} % 0.40 
\def \sfs {1} % 0.83
\begin{figure*}
\centering
% \captionsetup[subfigure]{aboveskip=0pt,belowskip=-7pt}
% \captionsetup[subfigure]{belowskip=-2pt}
\begin{subfigure}{\fs\textwidth}
\centering
\includegraphics[width=\sfs\textwidth]{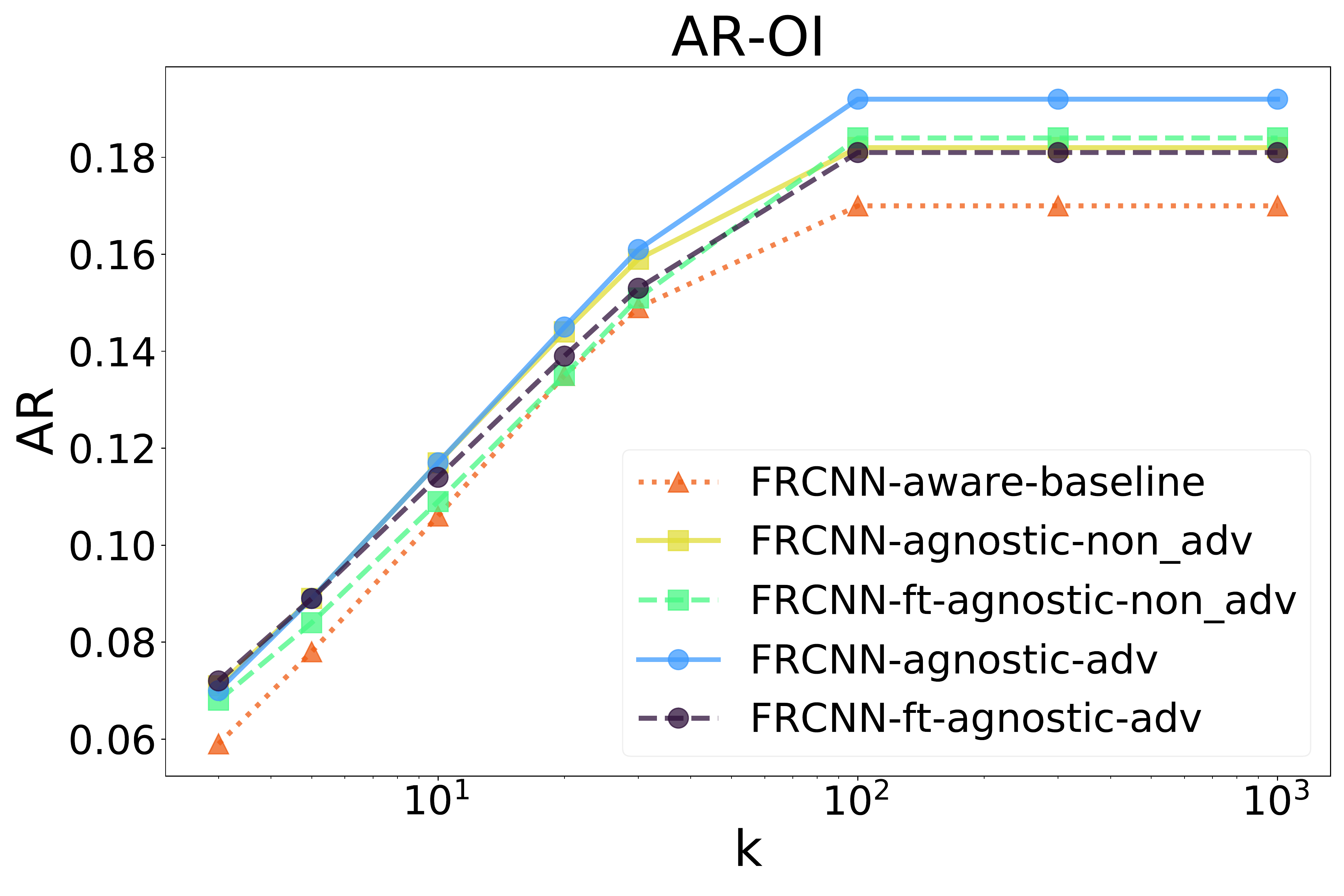}
\caption{Results of FRCNN models}
\end{subfigure}
\begin{subfigure}{\fs\textwidth}
\centering
\includegraphics[width=\sfs\textwidth]{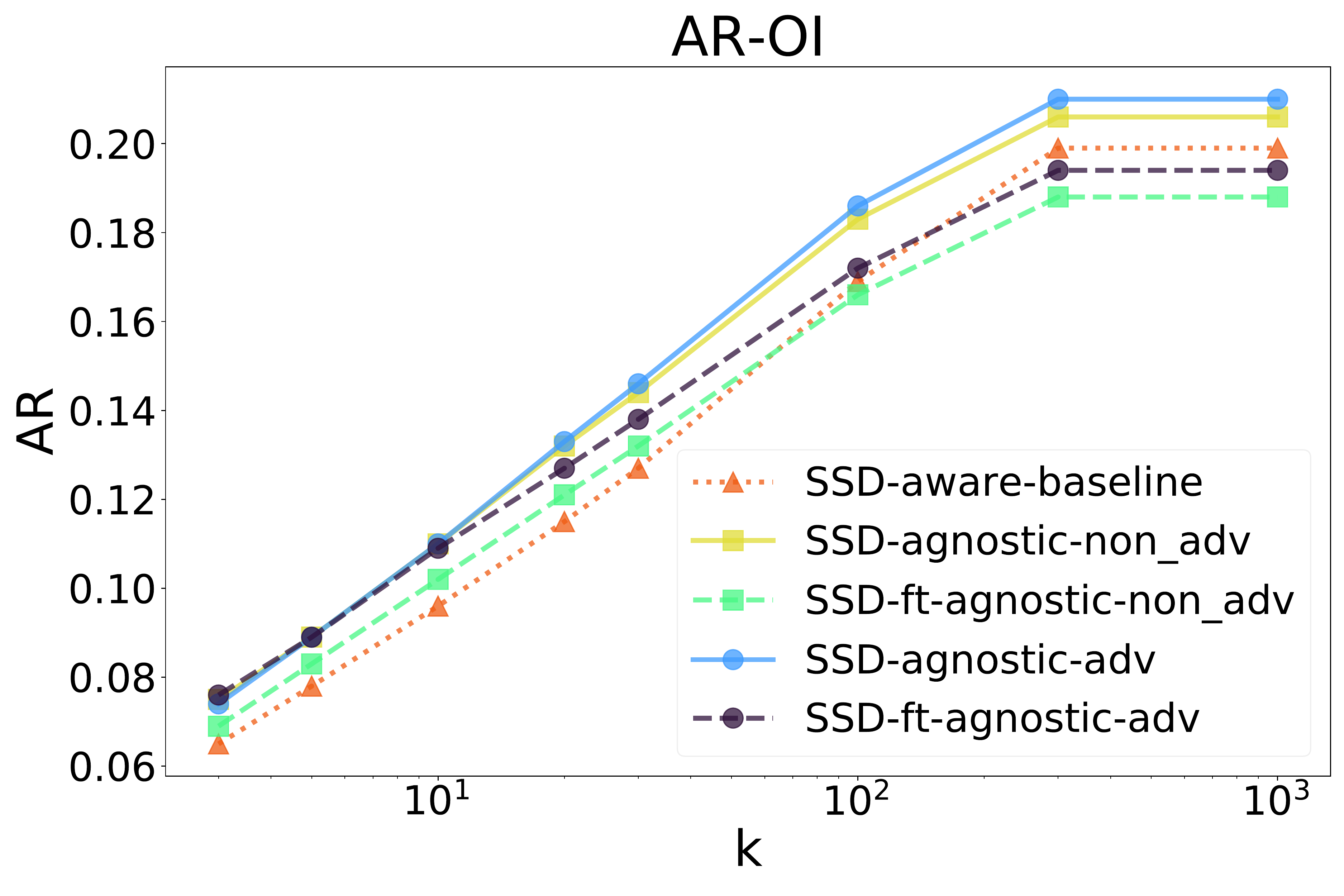}
\caption{Results of SSD models}
\end{subfigure}
\begin{subfigure}{\fs\textwidth}
\scriptsize
\centering
\vspace{-30pt}
\caption*{Terminology}
\setlength{\tabcolsep}{0.9em}
\begin{tabular}{c c}
\toprule
    \textbf{Token} & \textbf{Meaning} \\
    \cmidrule(r){1-1} \cmidrule(l){2-2}
    -aware- & Class-aware model \\
    -agnostic- & Class-agnostic model \\
    -ft- & Finetuned from -aware-baseline \\
    -non\_adv & Non-adversarial model \\
    -adv & Adversarially trained model \\
\bottomrule
\end{tabular}
\end{subfigure}
\caption{Generalization results of (a) FRCNN and (b) SSD models trained on the COCO dataset and evaluated on the non-overlapping classes in the Open Images dataset. Results show that class-agnostic models generalize better than the class-aware models, with the SSD-agnostic-adv and FRCNN-agnostic-adv models achieving the best recall.}
\vspace{-5pt}
\end{figure*}
}

\begin{table*}
\centering
\caption{\label{tab:objectnet}Downstream ObjectNet recognition results. Images are cropped using boxes predicted by the detectors and pretrained ImageNet models are used for classification from the cropped images. Averages of accuracies across classifiers are also reported. BO-acc is the accuracy when the predicted box with the highest IoU with the ground-truth box is used. Accuracies are also presented for top $M$ proposals as Acc@$M$. Results for uncropped images and those cropped using ground-truth boxes are provided for reference. ``-aw-'' and ``-ag-'' in the model name indicate whether the model is class-aware or -agnostic; ``-ad'' represents models trained adversarially. Results of other baseline detectors are provided in the supplementary material. Adversarially trained class-agnostic models achieve the best results.}
% Results of other baselines are provided in supplementary material.
% , ``-ft-'' tells whether the model was finetuned from the class-aware baseline,
\setlength{\tabcolsep}{0.6em} % for horizontal padding
\begin{tabular}{l l c c c c c c c c}
 \toprule
 & & \multicolumn{4}{c}{\textbf{Detectors trained on VOC}} & \multicolumn{4}{c}{\textbf{Detectors trained on COCO}} \\
 \cmidrule(lr){3-6} \cmidrule(lr){7-10}
 \textbf{Classifier} & \textbf{Detector} & \textbf{BO-acc} & \textbf{Acc$\text{@}$1} & \textbf{Acc$\text{@}$5} & \textbf{Acc$\text{@}$10} & \textbf{BO-acc} & \textbf{Acc$\text{@}$1} & \textbf{Acc$\text{@}$5} & \textbf{Acc$\text{@}$10} \\
\cmidrule(r){1-1} \cmidrule(lr){2-2} \cmidrule(lr){3-3} \cmidrule(lr){4-6}\cmidrule(lr){7-7} \cmidrule(lr){8-10}
%  None &  & \\
%  Ground-truth & & \\
%  \cmidrule(r){1-1} \cmidrule(lr){2-5} \cmidrule(lr){6-9}
\multirow{2}{*}{ResNet-152} & FRCNN-aw & 0.235 & 0.108 & 0.255 & 0.318 & 0.111 & 0.011 & 0.014 & 0.014 \\
  & FRCNN-ag-ad & \textbf{0.293} & \textbf{0.132} & \textbf{0.297} & \textbf{0.366} & \textbf{0.304} & \textbf{0.131} & \textbf{0.288} & \textbf{0.352} \\
 \cmidrule[0.1pt](r){1-2} \cmidrule[0.1pt](lr){3-6} \cmidrule[0.1pt](lr){7-10}
  \multirow{2}{*}{MobileNet-v2} & FRCNN-aw & 0.153 & 0.057 & \textbf{0.177} & 0.226 & 0.063 & 0.009 & 0.011 & 0.011 \\
    & FRCNN-ag-ad & \textbf{0.198} & \textbf{0.070} & \textbf{0.177} & \textbf{0.248} & \textbf{0.202} & \textbf{0.085} & \textbf{0.198} & \textbf{0.253} \\
\cmidrule[0.1pt](r){1-2} \cmidrule[0.1pt](lr){3-6} \cmidrule[0.1pt](lr){7-10}
  \multirow{2}{*}{Inception-v3} & FRCNN-aw & 0.194 & 0.072 & 0.223 & 0.288 & 0.094 & 0.011 & 0.013 & 0.014 \\
    & FRCNN-ag-ad & \textbf{0.249} & \textbf{0.086} & \textbf{0.224} & \textbf{0.310} & \textbf{0.258} & \textbf{0.106} & \textbf{0.255} & \textbf{0.320} \\
\cmidrule[0.5pt](r){1-2} \cmidrule[0.5pt](lr){3-6} \cmidrule[0.5pt](lr){7-10}
\multicolumn{1}{r}{\multirow{2}{*}{\textit{Avg. for}}} & FRCNN-aw & 0.194 & 0.079 & 0.218 & 0.277 & 0.089 & 0.010 & 0.013 & 0.013 \\
    \multicolumn{1}{r}{} & FRCNN-ag-ad & \textbf{0.247} & \textbf{0.096} & \textbf{0.233} & \textbf{0.308} & \textbf{0.255} & \textbf{0.107} & \textbf{0.247} & \textbf{0.308} \\ % ADD RELATIVE IMPROVEMENT
 \cmidrule[0.6pt]{1-10}
 \multirow{2}{*}{ResNet-152} & SSD-aw & 0.291 & 0.188 & 0.304 & 0.362 & 0.314 & 0.181 & 0.333 & 0.388 \\
 & SSD-ag-ad & \textbf{0.312} & \textbf{0.199} & \textbf{0.339} & \textbf{0.406} & \textbf{0.317} & \textbf{0.218} & \textbf{0.370} & \textbf{0.422} \\
 \cmidrule[0.1pt](r){1-2} \cmidrule[0.1pt](lr){3-6} \cmidrule[0.1pt](lr){7-10}
  \multirow{2}{*}{MobileNet-v2} & SSD-aw & 0.190 & 0.114 & 0.202 & 0.254 & 0.210 & 0.116 & 0.232 & 0.280 \\
    & SSD-ag-ad & \textbf{0.211} & \textbf{0.121} & \textbf{0.233} & \textbf{0.292} & \textbf{0.213} & \textbf{0.143} & \textbf{0.258} & \textbf{0.307} \\
\cmidrule[0.1pt](r){1-2} \cmidrule[0.1pt](lr){3-6} \cmidrule[0.1pt](lr){7-10}
  \multirow{2}{*}{Inception-v3} & SSD-aw & 0.247 & 0.156 & 0.270 & 0.332 & \textbf{0.270} & 0.148 & 0.301 & 0.361 \\
    & SSD-ag-ad & \textbf{0.267} & \textbf{0.165} & \textbf{0.308} & \textbf{0.383} & \textbf{0.270} & \textbf{0.182} & \textbf{0.331} & \textbf{0.390} \\
\cmidrule[0.5pt](r){1-2} \cmidrule[0.5pt](lr){3-6} \cmidrule[0.5pt](lr){7-10}
  \multicolumn{1}{r}{\multirow{2}{*}{\textit{Avg. for}}} & SSD-aw & 0.243 & 0.153 & 0.259 & 0.316 & 0.265 & 0.148 & 0.289 & 0.343 \\
   \multicolumn{1}{r}{} & SSD-ag-ad & \textbf{0.263} & \textbf{0.162} & \textbf{0.293} & \textbf{0.360} & \textbf{0.267} & \textbf{0.181} & \textbf{0.320} & \textbf{0.373} \\ % ADD RELATIVE IMPROVEMENT
 \cmidrule[0.75pt]{1-10}
 \multicolumn{9}{l}{Accuracy on full uncropped images} & 0.167\\
 \multicolumn{9}{l}{Accuracy on images cropped with ground-truth boxes (empirical upper bound of BO-Acc and Acc@1)} & 0.378 \\
 \bottomrule
%  \multicolumn{10}{c}{*Acc@5 is equivalent to ImageNet localization metric}
\end{tabular}
\vspace{-5pt}
\end{table*}

\subsection{Downstream Object Recognition}

We evaluate the object recognition performance for images cropped using bounding boxes predicted by the class-aware and class-agnostic models trained for the generalization experiments in \Cref{subsec:gen}. \Cref{tab:objectnet} summarizes the results of these experiments. Results show that the adversarially trained class-agnostic models perform better than the baselines in general on both Accuracy@$M$ and Best-overlap accuracy (as described in \Cref{subsec:downstream}), irrespective of the ImageNet classifier used. These results correlate with the generalization results presented in \Cref{subsec:results_gen}, indicating that detection models that generalize better to unseen object-types have higher downstream utility.

%% file: sections/06_conclusion.tex
\section{Conclusion and Future Work}
\label{sec:conclusion}

Conventional class-aware objection detection models perform well at detecting ``objects of interest'' that belong to known objects types seen during training. However, these models do not generalize well to unseen object-types, limiting their incorporation in real-world applications. In this work, we have formulated a novel task of class-agnostic object detection, where the goal is to detect objects of known \emph{and} unknown types irrespective of their category. Furthermore, we have presented training and evaluation protocols for benchmarking models and advancing future research in this direction. These include two sets of experiments -- (1) one for generalization to new object-types and (2) another for downstream utility in terms of object recognition. Finally, we have presented a few intuitive baselines and proposed a new adversarially trained model that penalizes the objective if the learned representations encode object-type. Results show that the proposed adversarial class-agnostic model outperforms the baselines on both generalization and downstream-utility experiments, for both one-stage (SSD) and two-stage (Faster R-CNN) detectors.

This work serves to establish a new direction of research in object detection. As such, the possibilities for future research are endless, including but not limited to --- (1) further refinement of the problem formulation, (2) improved as well as additional training and evaluation protocols, and (3) novel methods for class-agnostic object detection.